\documentclass{article}

\usepackage{microtype}
\usepackage{graphicx}
\usepackage{subfigure}
\usepackage{booktabs} 
\usepackage{amssymb}
\usepackage{amsmath}
\usepackage{amssymb}
\usepackage{enumitem}
\usepackage{setspace}

\graphicspath{{final_images/}}
\newtheorem{theorem}{Theorem}
\newtheorem{definition}{Definition}

\usepackage{hyperref}

\usepackage[accepted]{icml2019}

\icmltitlerunning{Rethinking Lossy Compression: {T}he Rate-Distortion-Perception Tradeoff}

\begin{document}

\twocolumn[
\icmltitle{Rethinking Lossy Compression: {T}he Rate-Distortion-Perception Tradeoff}




\begin{icmlauthorlist}
\icmlauthor{Yochai Blau}{tech}
\icmlauthor{Tomer Michaeli}{tech}
\end{icmlauthorlist}

\icmlaffiliation{tech}{Technion--Israel Institute of Technology, Haifa, Israel}

\icmlcorrespondingauthor{Yochai Blau}{yochai@campus.technion.ac.il}
\icmlcorrespondingauthor{Tomer Michaeli}{tomer.m@ee.technion.ac.il}


\vskip 0.3in
]



\printAffiliationsAndNotice{}  

\begin{abstract}
Lossy compression algorithms are typically designed and analyzed through the lens of Shannon's rate-distortion theory, where the goal is to achieve the lowest possible distortion (e.g., low MSE or high SSIM) at any given bit rate. However, in recent years, it has become increasingly accepted that ``low distortion'' is not a synonym for ``high perceptual quality'', and in fact optimization of one often comes at the expense of the other. In light of this understanding, it is natural to seek for a generalization of rate-distortion theory which takes perceptual quality into account. In this paper, we adopt the mathematical definition of perceptual quality recently proposed by Blau \& Michaeli \yrcite{blau2018perception}, and use it to study the three-way tradeoff between rate, distortion, and perception. We show that restricting the perceptual quality to be high, generally leads to an elevation of the rate-distortion curve, thus necessitating a sacrifice in either rate or distortion. We prove several fundamental properties of this triple-tradeoff, calculate it in closed form for a Bernoulli source, and illustrate it visually on a toy MNIST example.
\end{abstract}
\section{Introduction}
Lossy compression techniques are ubiquitous in the modern-day digital world, and are regularly used for communicating and storing images, video and audio. 
In recent years, lossy compression is seeing a surge of research, due in part to the advancements in deep learning and their application in this domain \cite{toderici2016variable,toderici2017full,balle2016end,balle2017end,balle2018variational,agustsson2017soft,agustsson2018generative,rippel2017real,minnen2018joint,li2018learning,mentzer2018conditional,johnston2018improved,galteri2017deep,tschannen2018deep,santurkar2018generative,rott2018deformation}. The theoretical foundations of lossy compression are rooted in Shannon's seminal work on rate-distortion theory \cite{shannon1959coding}, which analyzes the fundamental tradeoff between the bit rate used for representing data, and the distortion incurred when reconstructing the data from its compressed representation \cite{cover2012elements}.

The premise in rate-distortion theory is that reduced distortion is a desired property. However, recent works demonstrate that minimizing distortion alone does not necessarily drive the decoded signals to have good perceptual quality. For example, incorporating generative adversarial type losses has been shown to lead to significantly better perceptual quality, but at the cost of \emph{increased} distortion \cite{tschannen2018deep, agustsson2018generative, santurkar2018generative}. This behavior has also been studied in the context of signal restoration \cite{blau2018perception}, where it was shown that minimizing distortion causes the distribution of restored signals to deviate from that of the ground-truth signals (indicating worse perceptual quality). In light of this understanding, it is natural to seek for a generalized rate-distortion theory, which also accounts for perception. In particular, it is of key importance to understand how the best achievable rate depends not only on the distortion, but also on the perceptual quality of the algorithm. A preliminary attempt to incorporate perceptual quality into rate-distortion theory was briefly reported in \cite{matsumoto2018introducing,matsumoto2018rate}. Yet, no theoretical characterization nor practical demonstration of its effect on the rate-distortion tradeoff was presented. 

\begin{figure}[t]
	\vskip 0.05in
	\begin{center}
		\centerline{\includegraphics[width=0.9\columnwidth]{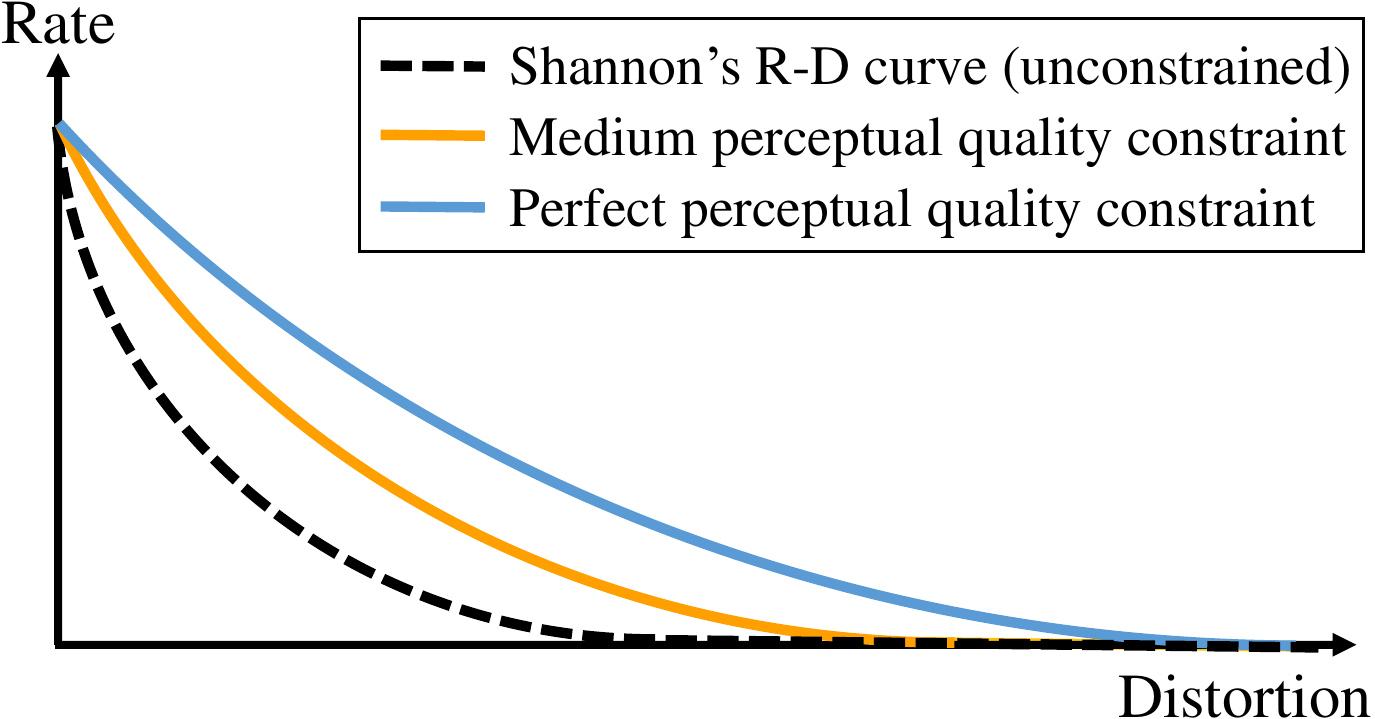}}
		\caption{\textbf{The rate-distortion function under a perceptual quality constraint.} When perceptual quality is unconstrained, the tradeoff is characterized by Shannon's classic rate-distortion function (black line). However, as the constraint on perceptual quality is tightened to ensure perceptually pleasing reconstructions, the function elevates (colored lines). Thus, the improvement in perceptual quality comes at the cost of a higher rate and\slash or distortion.}
		\label{fig:schematic}
	\end{center}
	\vskip -0.25in
\end{figure}

In this paper, we adopt the mathematical definition of perceptual quality used in \cite{blau2018perception}, 
and prove that there is a triple tradeoff between rate, distortion \emph{and} perception. Our key observation is that the rate-distortion function elevates as the perceptual quality is enforced to be higher (see Fig.~\ref{fig:schematic}). In other words, to obtain good perceptual quality, it is necessary to make a sacrifice in either the distortion or the rate of the algorithm.

Our analysis is based on the definition of a \emph{rate-distortion-perception function} $R(D,P)$, which characterizes the minimal achievable rate $R$ for any given distortion $D$ and perception index $P$.
We begin by deriving a closed form for this function in the classical case study of a Bernoulli source, a simple example which nonetheless nicely illustrates the typical behavior of the tradeoff. We then prove several general properties of $R(D,P)$, showing that it is monotone and convex for any full-reference distortion measure (under minor assumptions), and that there is a range of $P$ values for which it necessarily does not coincide with the traditional rate-distortion function. For the specific case of the squared-error distortion, we also provide an upper bound on the increase in distortion that has to be incurred in order to achieve perfect perceptual quality, at any given rate.

Our observations have important implications for the design and evaluation of practical compression methods. In particular, they suggest that comparing between algorithms only in terms of their rate-distortion curves can be misleading. 
We demonstrate this in the context of image compression using a toy MNIST example, by systematically exploring the visual effect of improvement in each of the three properties (rate, distortion, perception) on the expense of the others. 
We do this by training an encoder-decoder net utilizing a generative model, similarly to \cite{tschannen2018deep, agustsson2018generative}. As we show, the phenomena we discuss are dominant at low bit rates, where the classical approach of optimizing distortion alone leads to unacceptable perceptual quality. 
This is perhaps not surprising when using the MSE distortion, which is known to be inconsistent with human perception. But our theory shows that \emph{every} distortion measure (excluding pathological cases) must have a tradeoff with perceptual quality. This includes e.g., the popular SSIM\slash MS-SSIM \cite{wang2003multiscale,wang2004image}, the $L_2$ distance between deep features \cite{johnson2016perceptual}, and any other full reference criterion. To illustrate this, we repeat our toy experiment with the distortion measure of \cite{johnson2016perceptual}, which has been used as a means for enhancing perceptual quality in low-level vision tasks \cite{ledig2017photo}. As we show, minimizing this distortion \emph{does not} lead to good perceptual quality at low bit rates, just like our theory predicts. Moreover, when enforcing high perceptual quality, this distortion rather increases. 

\section{Background}

\subsection{Rate-Distortion Theory}
\begin{figure}[t]
	\begin{center}
		\centerline{\includegraphics[width=0.9\columnwidth]{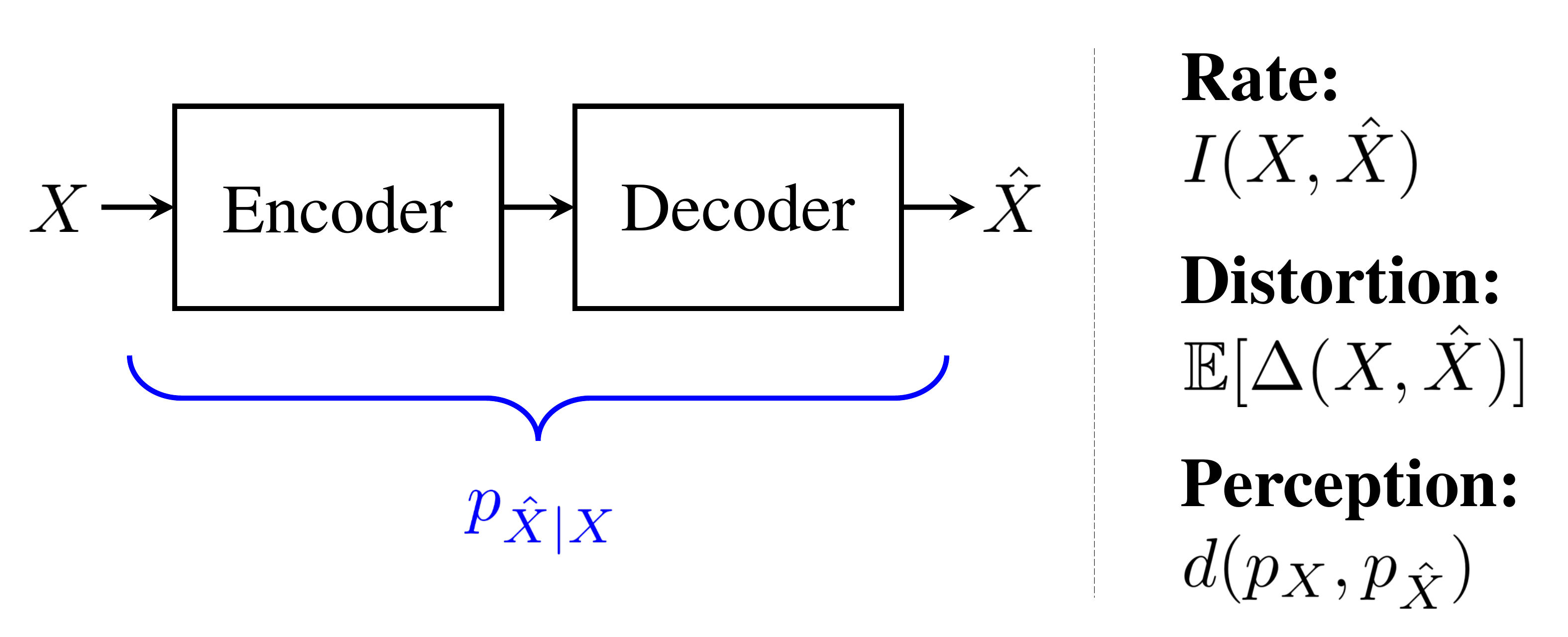}}
		\caption{\textbf{Lossy compression.} A source signal $X\!\sim\! p_X$ is mapped into a coded sequence by an encoder and back into an estimated signal $\hat{X}$ by the decoder. Three desired properties are: (i) the coded sequence be compact (low bit rate); (ii) the reconstruction $\hat{X}$ be similar to the source $X$ on average (low distortion); (iii)~the distribution $p_{\hat{X}}$ be similar to $p_X$, so that decoded signals are perceived as genuine source signals (good perceptual quality).}
		\label{fig:scheme}
	\end{center}
	\vskip -0.25in
\end{figure}

Rate-distortion theory analyzes the fundamental tradeoff between the rate (bits per sample) used for representing samples from a data source $X\sim p_X$, and the expected distortion incurred in decoding those samples from their compressed representations. Formally, the relation between the input $X$ and output $\hat{X}$ of an encoder-decoder pair, is a (possibly stochastic) mapping defined by some conditional distribution $p_{\hat{X}|X}$, as visualized in Fig.~\ref{fig:scheme}. The expected distortion of the decoded signals is thus defined as
\begin{align}
    \mathbb{E}[\Delta(X,\hat{X})],
\end{align}
where the expectation is with respect to the joint distribution $p_{X,\hat{X}}=p_{\hat{X}|X}p_X$, and $\Delta:\mathcal{X} \times \hat{\mathcal{X}} \rightarrow \mathbb{R}^+$ is any full-reference distortion measure such that $\Delta(x,\hat{x})=0$ if and only if $x=\hat{x}$  (e.g., squared error, $L_2$ distance between deep features \cite{johnson2016perceptual,zhang2018unreasonable}, SSIM\slash MS-SSIM\footnote{Measures like SSIM, which quantify similarity rather than dissimilarity and are not necessarily positive, need to be negated and shifted to qualify as valid distortion measures.} \cite{wang2003multiscale,wang2004image}, PESQ \cite{rix2001perceptual}, etc.).

A key result in rate-distortion theory states that for an iid source $X$, if the expected distortion is bounded by $D$, then the lowest achievable rate $R$ is characterized by the (information) rate-distortion function
\begin{equation}\label{eq:RD}
R(D) = \min_{p_{\hat{X}|X}} \, I(X,\hat{X}) \quad \textrm{s.t.} \quad \mathbb{E}[\Delta(X,\hat{X})] \le D,
\end{equation}
where $I$ denotes mutual information \cite{cover2012elements}. 
Closed form expressions for the rate-distortion function $R(D)$ are known for only a few source distributions and under quite simple distortion measures (e.g., squared error or Hamming distance). However several general properties of this function are known, 
including that it is always monotonically non-increasing and convex. 

\subsection{Perceptual Quality}\label{sec:PD}
The perceptual quality of an output sample $\hat{x}$ refers to the extent to which it is perceived by humans as a valid (natural) sample, regardless of its similarity to the input $x$. In various domains, perceptual quality has been associated with the deviation of the distribution $p_{\hat{X}}$ of output signals from the distribution $p_X$ of natural signals, which, as discussed in \cite{blau2018perception}, is linked to the common practice of quantifying perceptual quality via real-vs.-fake user studies \cite{isola2017image,salimans2016improved,zhang2016colorful,denton2015deep}. In particular, deviation from natural scene statistics is the basis for many no-reference image quality measures \cite{mittal2013making,mittal2012no,wang2005reduced}, which have been shown to correlate well with human opinion scores. It is also the principle underlying GAN-based image restoration schemes, which achieve enhanced perceptual quality by directly minimizing some divergence $d(p_X,p_{\hat{X}})$  \cite{ledig2017photo,pathak2016context,isola2017image,wang2018esrgan}. Based on these works, and following \cite{blau2018perception}, we define the perceptual quality index (lower is better) of an algorithm as
\begin{equation}\label{eq:perceptualQuality}
d(p_X,p_{\hat{X}}),
\end{equation}
where $d(\cdot,\cdot)$ is some divergence between distributions\footnote{We assume that $d(p,q)\ge0, d(p,q)=0\Leftrightarrow p=q$.} (e.g., Kulback-Leibler, Wasserstein, etc.). Note that the divergence function $d(\cdot,\cdot)$ which best relates to human perception is a subject of ongoing research. Yet, our results below hold for (nearly) any divergence.

Obviously, perceptual quality, as defined above, is very different from distortion. In particular, minimizing the perceptual quality index does not necessarily lead to low distortion. For example, if the decoder disregards the input, and outputs random samples from the source distribution $p_{X}$, it will achieve perfect perceptual quality but very poor distortion. It turns out that this is true also in the other direction. That is, minimizing distortion does not necessarily lead to good perceptual quality. This observation has been studied in \cite{blau2018perception} in the specific context of signal restoration (e.g.~denoising, super-resolution). In particular, perception and distortion are fundamentally at odds with each other (for non-invertible degradations), in the sense that optimizing one always comes on the expense of the other. This behavior, coined \emph{the perception-distortion tradeoff}, was shown to hold true for \emph{any} distortion measure.
\section{The Rate-Distortion-Perception Tradeoff}
Since both perceptual quality and distortion are typically important, here we extend the rate-distortion function \eqref{eq:RD} to take into account the perception index\footnote{Similarly to \eqref{eq:RD}, $R(D,P)$ in \eqref{eq:RDP} lower bounds the best achievable rate for an iid source (see Supplementary Material). We do not prove achievability of $R(D,P)$ in general. However for the MSE distortion, we show an achievable upper bound (see Theorem~\ref{thm:bound}).} \eqref{eq:perceptualQuality}.

\begin{definition}
The (information) rate-distortion-perception function is defined as
\begin{align}\label{eq:RDP}
R(D,P) = &\min_{p_{\hat{X}|X}} \,\, I(X,\hat{X}) \nonumber \\ 
&\,\, \text{\emph{s.t.}} \,\, \mathbb{E}[\Delta(X,\hat{X})] \le D,\,\, d(p_X,p_{\hat{X}}) \le P.
\end{align}
\end{definition}

Unfortunately, closed form solutions for \eqref{eq:RDP} are even harder to obtain than for \eqref{eq:RD}. Yet, one notable exception is the classical case study of a binary source, as we show next. While of limited applicability, this example illustrates the typical behavior of \eqref{eq:RDP}, which we analyze in Sec.~\ref{sec:theoreticalProperties}.

\subsection{Bernoulli Source}\label{sec:Bernoulli}
Consider the problem of encoding a binary source $X \sim \text{Bern}(p)$, where the decoder's output $\hat{X}$ is also constrained to be binary. Let us take the distortion measure $\Delta(\cdot,\cdot)$ to be the Hamming distance, and the perception index\footnote{The term ``perception'' is somewhat inappropriate for a Bernoulli source, as it is not \emph{perceived} by humans (contrary to images, audio). Yet, we keep this terminology here for consistency.} to be the total-variation (TV) distance $d_{\text{TV}}(\cdot,\cdot)$. Without loss of generality, we assume that $p \le \tfrac{1}{2}$. 
When perception is not constrained (i.e.,~$P=\infty$), the solution to \eqref{eq:RDP} reduces to the rate-distortion function \eqref{eq:RD} of a binary source, which is known to be given by
\begin{equation}\label{eq:RD_Bern}
R(D,\infty) =
\begin{cases}
H_b(p) - H_b(D) \quad \quad & D\in[0,p)\\
0 & D \in[p,\infty)
\end{cases}
\end{equation}
where $H_b(\alpha)$ is the entropy of a Bernoulli random variable with probability $\alpha$ \cite{cover2012elements}. 

In the Supplementary Material, we derive the solution for arbitrary $P$. It turns out that as long as the perceptual quality constraint is sufficiently loose, the solution remains the same. However, when $P \le p$, the perception constraint in \eqref{eq:RDP} becomes active whenever the distortion constraint is loose enough, from which point the function $R(\cdot,P)$ departs from $R(\cdot,\infty)$. Specifically, for $P \le p$, we have
\begin{align}\label{eq:Bern_RDP}
&R(D,P) =  \\ \nonumber
& \!\!\begin{cases}
H_b(p) - H_b(D) & \!\!\!D\!\in\!\mathcal{S}_1  \\
2H_b(p)\!+\!H_b(p\!-\!P)\!-\!H_t(\tfrac{D-P}{2},p)\!-\!H_t(\tfrac{D+P}{2},q)  & \!\!\!D\!\in\!\mathcal{S}_2\\
0 & \!\!\!D\!\in\!\mathcal{S}_3
\end{cases}
\end{align}
where $q = 1 - p$ and $H_t(\alpha,\beta)$ denotes the entropy of a ternary random variable with probabilities $\alpha$, $\beta$, $1-\alpha-\beta$. Here, $\mathcal{S}_1=[0,D_1)$, $\mathcal{S}_2=[D_1,D_2)$, and $\mathcal{S}_3=[D_2,\infty)$, where $D_1=\tfrac{P}{1-2(p-P)}$ and $D_2=2pq-(q-p)P$.

\begin{figure}[t]
	\vskip 0.05in
	\begin{center}
		\centerline{\includegraphics[width=\columnwidth]{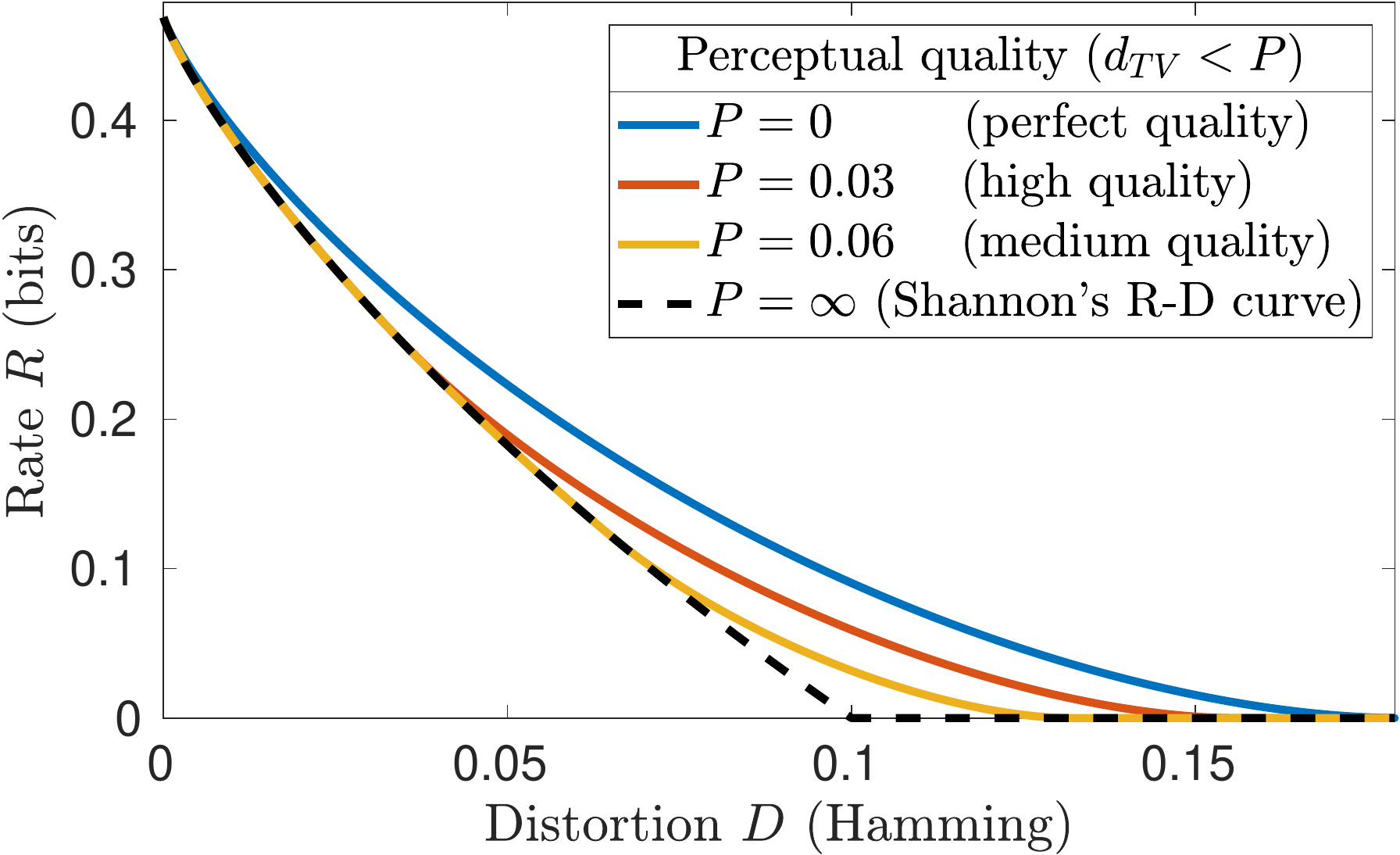}}
		\caption{\textbf{Perception constrained rate-distortion curves for a Bernoulli source.} Shannon's rate-distortion function (dashed curve) characterizes the best achievable rate under any prescribed distortion level, yet does not ensure good perceptual quality. When constraining the perceptual quality index $d_{\text{TV}}(p_X,p_{\hat{X}})$ to be low (good quality), the rate-distortion function elevates (solid curves). This indicates that good perceptual quality must come at the cost of a higher rate and\slash or a higher distortion. Here $X\sim\text{Bern}(\tfrac{1}{10})$.}
		\label{fig:Bern_RD}
	\end{center}
	\vskip -0.25in
\end{figure}

Figure~\ref{fig:Bern_RD} plots $R(D,P)$ as a function of $D$ for several values of $P$. As can be seen, at $D=0$, all the curves merge. This is because at this point $\hat{X}=X$ (lossless compression), so that $p_{\hat{X}}=p_X$, and thus the perceptual quality is perfect. Yet, as the allowed distortion $D$ grows larger, the curves depart. This illustrates that achieving the classical rate-distortion curve (black dashed line) does not generally lead to good perceptual quality. The more stringent our prescribed perceptual quality constraint (lower $P$), the more the rate-distortion curve elevates (colored curves). In particular, the tradeoff becomes severe at the low bit rate regime, where good perceptual quality comes at the cost of a significantly higher distortion and\slash or bit rate. Notice that it is possible to achieve \emph{perfect} perceptual quality at every rate (blue curve) by compromising the distortion to some extent. In Sec.~\ref{sec:theoreticalProperties} we provide an upper-bound on the increase in distortion required for obtaining perfect perceptual quality.

\begin{figure*}[!t]
	\vskip 0.05in
	\centering
	\subfigure[]{%
		\includegraphics[width=.29\textwidth]{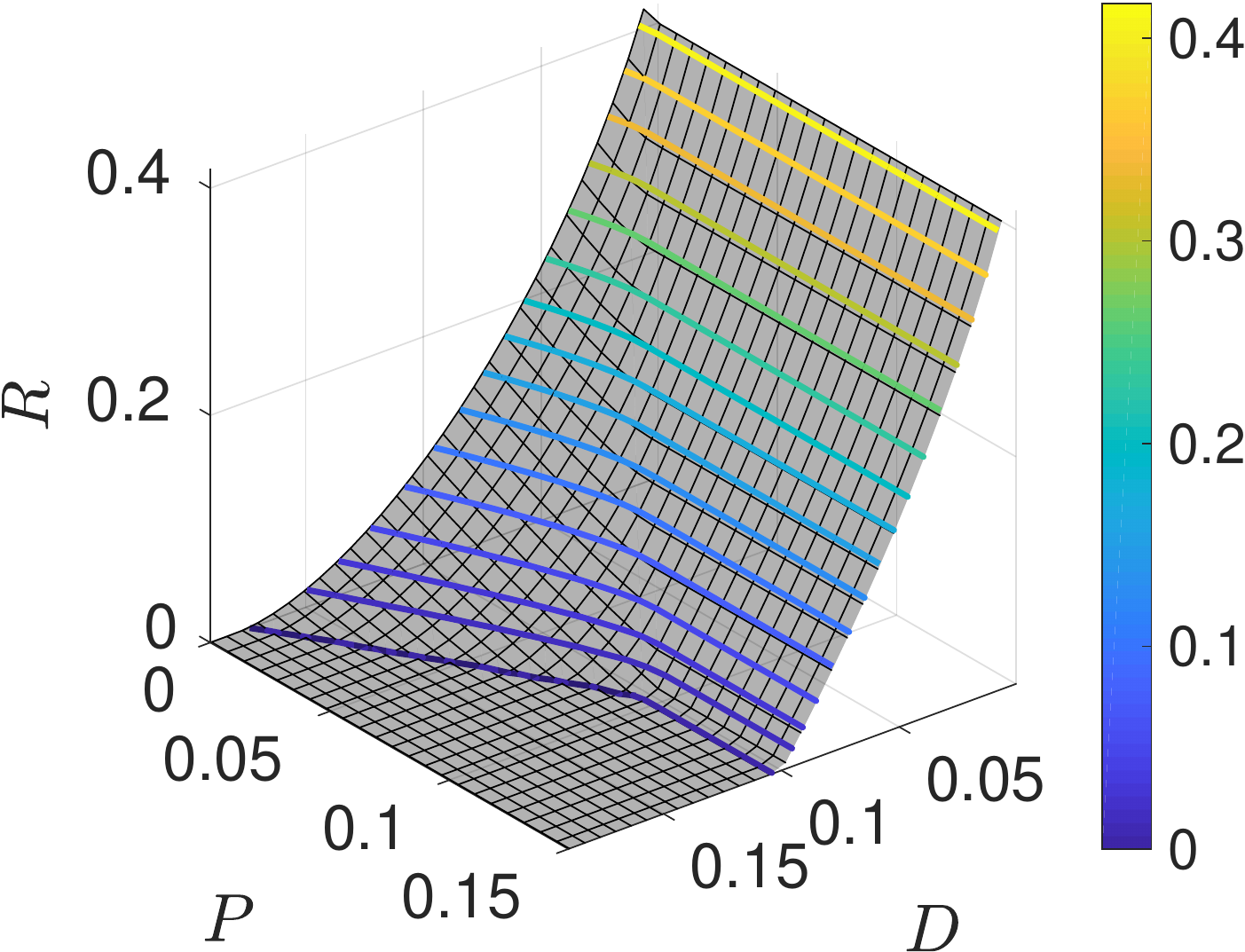} \label{fig:Bern_1} 
	} 
	\quad \quad
	\subfigure[]{%
		\includegraphics[width=.32\textwidth]{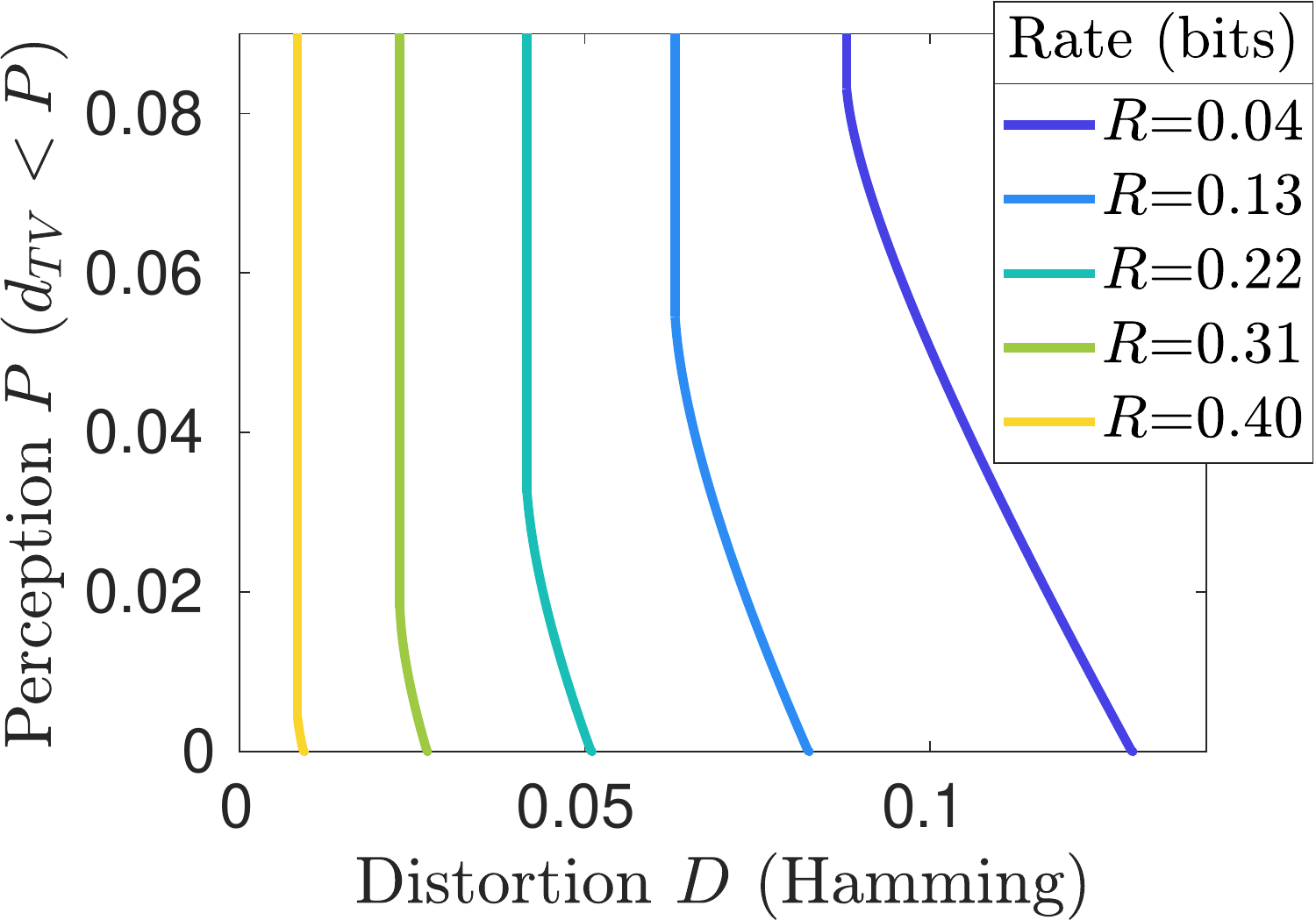}
		\label{fig:Bern_2} 
	} 
	\quad \quad
	\subfigure[]{%
		\includegraphics[width=.28\textwidth]{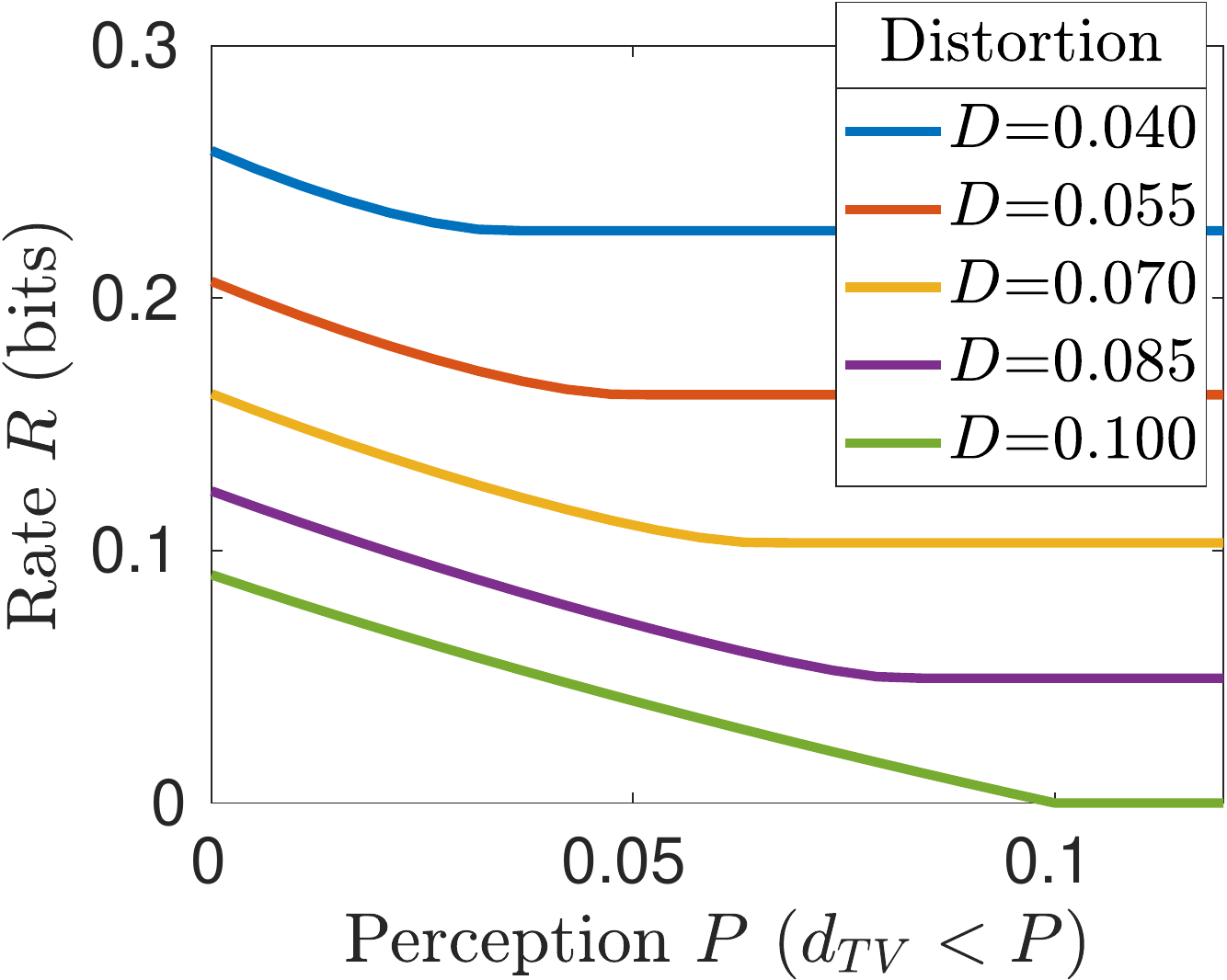}
		\label{fig:Bern_3} 
	} 
	\vskip -0.05in
	\caption{\textbf{The rate-distortion-perception function of a Bernoulli source.} (a) Equi-rate level sets depicted on the rate-distortion-perception function $R(D,P)$. At low bit-rates, the equi-rate lines curve substantially when approaching $P=0$, displaying the increasing tradeoff between distortion and perceptual quality. (b) Cross sections of $R(D,P)$ along perception-distortion planes. Notice the tradeoff between perceptual quality and distortion, which becomes stronger at low bit-rates. (c) Cross sections of $R(D,P)$ along rate-perception planes. Note that at \emph{constant} distortion, the perceptual quality can be improved by increasing the rate.}
	\label{fig:Bern_figs}
\end{figure*}

While Fig.~\ref{fig:Bern_RD} displays cross-sections of $R(D,P)$ along rate-distortion planes, in Fig.~\ref{fig:Bern_figs} we plot $R(D,P)$ as a surface in 3 dimensions, as well as its cross-sections along the other planes. The equi-rate level sets shown on the surface in Fig.~\ref{fig:Bern_1}, provide another visualization for the phenomenon described above. That is, at high bit-rates, it is possible to achieve good perceptual quality (low $P$) without a significant sacrifice in the distortion $D$. However, as the bit-rate becomes lower, the equi-rate level sets substantially curve towards the low $P$ values, illuminating the exacerbation in the tradeoff between distortion and perception in this regime. Figure~\ref{fig:Bern_2} provides an additional viewpoint, by showing perception-distortion curves for different bit rates. Notice again that the tradeoff between distortion and perceptual quality becomes stronger at low bit-rates. Finally, Fig.~\ref{fig:Bern_3} shows the somewhat counter-intuitive tradeoff between rate and perceptual quality as a function of distortion. Specifically, we see that at every \emph{constant} distortion level, the perceptual quality can be improved by increasing the rate.

\subsection{Theoretical Properties}\label{sec:theoreticalProperties}
For general source distributions, it is usually impossible to solve \eqref{eq:RDP} analytically. However, it turns out that the behavior we saw for a Bernoulli source is quite typical. We next prove several general properties of the function \eqref{eq:RDP}, which hold under rather mild assumptions. Specifically, we assume:
    
    \textbf{A1} The divergence $d(\cdot,\cdot)$ in \eqref{eq:RDP} is convex in its second argument. That is, for any $\lambda \in [0,1]$ and for any three distributions $p_0,q_1,q_2$,
    \begin{equation}
d(p_0,\lambda q_1 + (1-\lambda) q_2) \le \lambda d(p_0,q_1) + (1-\lambda) d(p_0,q_2).
\end{equation}
    \textbf{A2} The function $k(z)=\mathbb{E}_{X \sim p_X}[\Delta(X,z)]$ is not constant\footnote{In fact, we only need the weaker condition that $k(z)$ do not attain its minimum over the entire support of $p_X$.} over the entire support of $p_X$.

Assumption \textbf{A1} is not very limiting. For instance, any $f$-divergence (e.g. KL, TV, Hellinger, $\mathcal{X}^2$) as well as the Renyi divergence, satisfies this assumption \cite{csiszar2004information,van2014renyi}. Assumption \textbf{A2} holds in any setting where the mean distance between a ``valid'' signal $z$ and all other ``valid'' signals is not constant\footnote{A valid signal is any $x : p_X(x)>0$. Also, we use ``distance'' here for clarity, although $\Delta(\cdot,\cdot)$ is not necessarily a metric.}. In particular, it holds for any distortion function $\Delta(\cdot,\cdot)$ with a \emph{unique} minimizer, such as the squared-error distortion and the SSIM index (under some assumptions) \cite{brunet2012study}. Using these assumptions, we are able to qualitatively characterize the general shape of the function $R(D,P)$.
\begin{theorem}\label{thm:convexity}
	The rate-distortion-perception function \eqref{eq:RDP}:
	\vspace{-0.4cm}
	\begin{enumerate}[noitemsep]
		\item is monotonically non-increasing in $D$ and $P$;
		\item is convex if \textbf{A1} holds;
		\item satisfies $R(\cdot,0) \ne R(\cdot,\infty)$ if \textbf{A2} holds.
	\end{enumerate}
\end{theorem}

The proof of Theorem \ref{thm:convexity} can be found in the Supplementary Material. Note that when assumption \textbf{A2} holds, properties 1 and 3 indicate that there exists some $D_0$ for which $R(D_0,0) > R(D_0,\infty)$, showing that the rate-distortion curve necessarily elevates when constraining for perfect perceptual quality. In any case, assumption \textbf{A2} is a \emph{sufficient} condition for property 3, so that even if it does not hold, this does not necessarily imply that $R(\cdot,0) = R(\cdot,\infty)$.

How much does the rate-distortion curve elevate when constraining for \emph{perfect} perceptual quality? The next theorem upper-bounds this elevation for the MSE distortion (see proof in the Supplementary Material).

\begin{theorem}\label{thm:bound}
	When using the squared-error distortion, the function $R(\cdot,0)$ (rate-distortion at perfect perceptual quality) is bounded by
	\begin{equation}
	R(D,0) \le R(\tfrac{1}{2}D,\infty).
	\end{equation}
\end{theorem}

\begin{figure}[t]
	\vskip 0.05in
	\begin{center}
		\centerline{\includegraphics[width=0.9\columnwidth]{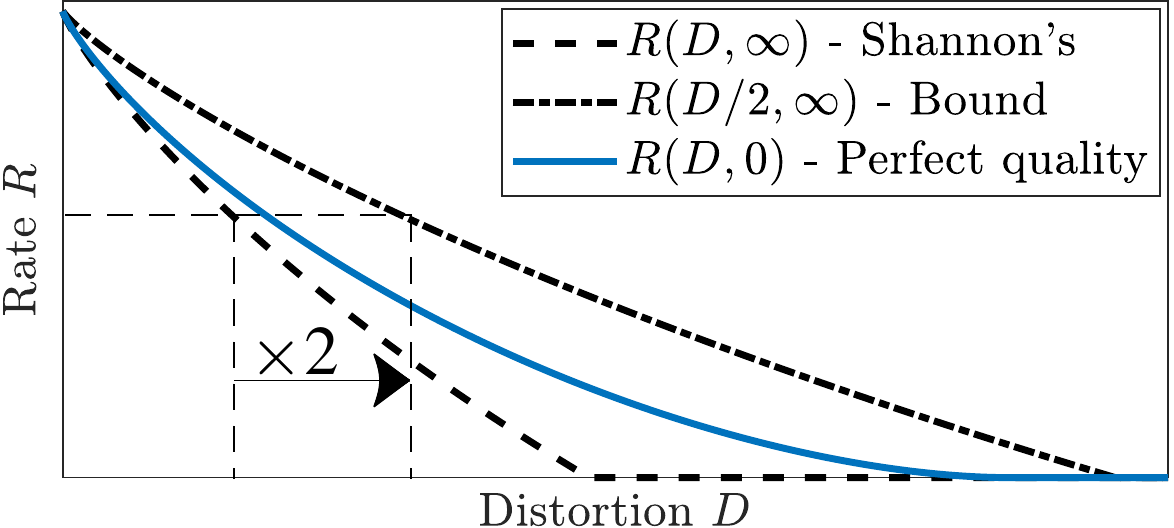}}
		\caption{\textbf{Illustration of Theorem \ref{thm:bound}.} When using the MSE distortion, the rate-distortion curve for compression with perfect perceptual quality (blue) is higher than Shannon's rate-distortion function (black dashed line) but is necessarily lower than the $2\times $ scaled version of Shannon's function (dotted line).}
		\label{fig:thm_2}
	\end{center}
	\vskip -0.25in
\end{figure}

Theorem \ref{thm:bound} shows that it is possible to attain perfect perceptual quality without increasing the rate, by sacrificing no more than a $2$-fold increase in the mean squared-error (MSE). More specifically, attaining perfect perceptual quality at distortion $D$ does not require a higher bit rate than that necessary for compression at distortion $\frac{1}{2}D$ with no perceptual quality constraint. This is illustrated in Fig.~\ref{fig:thm_2}, where the perfect-quality curve $R(\cdot,0)$ shown in blue is bounded by the scaled version of Shannon's unconstrained quality curve $R(\cdot,\infty)$ shown as a black dashed line. In image restoration scenarios, such a $2$-fold increase in the MSE ($3$dB decrease in PSNR) has been shown to enable a substantial improvement in perceptual quality by practical algorithms \cite{blau2018pirm,ledig2017photo}. Note that this bound is generally not tight. Thus, in some settings, perfect perceptual quality can be obtained with an even smaller increase in distortion. 
\section{Experimental Illustration}\label{sec:exp}
We now turn to demonstrate the visual implications of the rate-distortion-perception tradeoff in lossy image compression on a toy MNIST example. We make no attempt to propose a new state-of-the-art compression method. Our sole goal is to systematically explore the effect of the balance between rate, distortion, and perception. To this end, we utilize a net-based encoder-decoder pair trained in an end-to-end fashion, similarly to recent works. By tuning the influence of each of the different terms of the loss, we can easily control the balance between these three quantities.

\begin{figure*}[t]
	\vskip 0.05in
	\begin{center}
		\centerline{\includegraphics[width=0.95\textwidth]{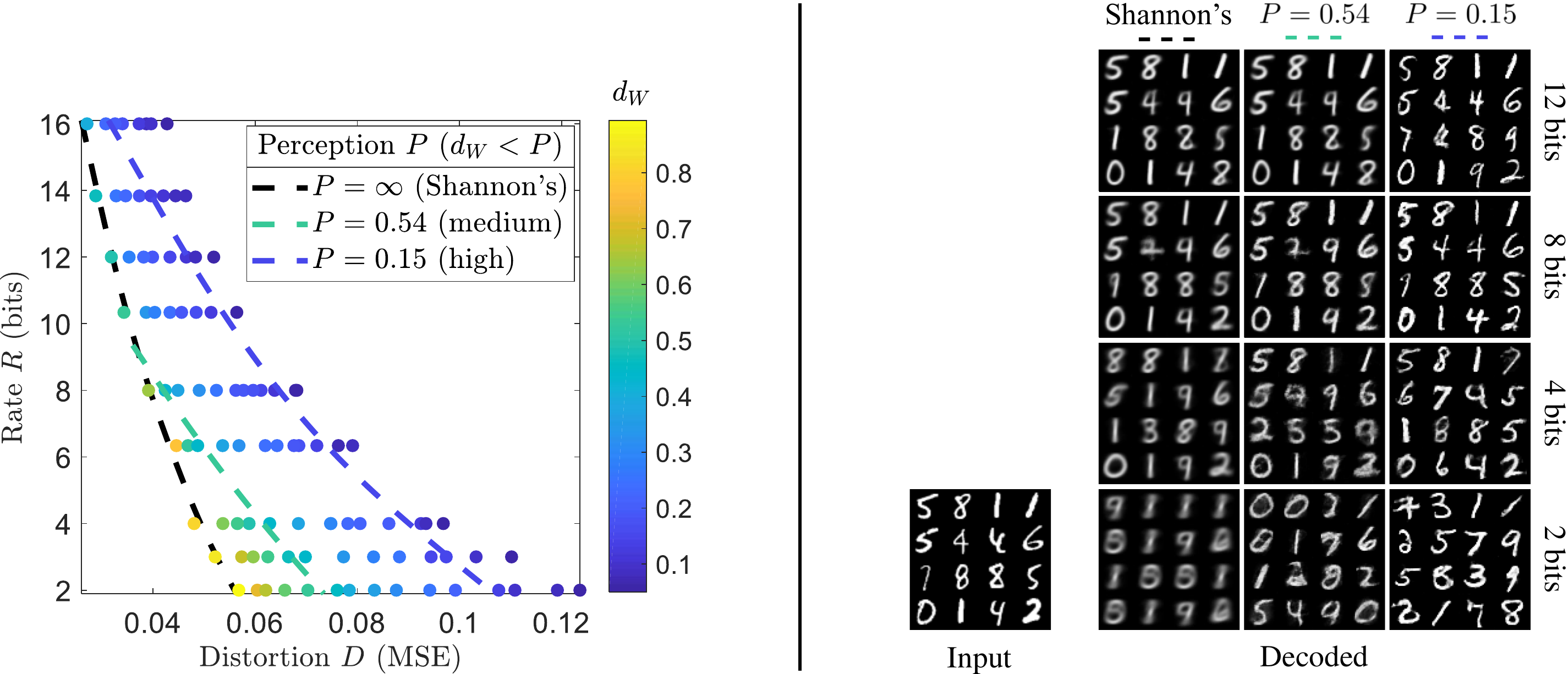}}
		\caption{\textbf{Perceptual lossy compression of MNIST digits.} \emph{Left}: Shannon's rate-distortion curve (black) describes the lowest possible rate (bits per digit) as a function of distortion, but leads to low perceptual quality (high $d_{\text{W}}$ values), especially at low rates. When constraining the perceptual quality to be good (low $P$ values), the rate-distortion curve elevates, indicating that this comes at the cost of a higher rate and\slash or distortion. \emph{Right}: Encoder-decoder outputs along Shannon's rate-distortion curve and along two equi-perceptual-quality curves. As the rate decreases, the perceptual quality along Shannon's curve degrades significantly. This is avoided when constraining the perceptual quality, which results in visually pleasing reconstructions, even at extremely low bit-rates. Notice that increased perceptually quality does not imply increased accuracy, as most reconstructions fail to preserve the digits' identities at a 2-bit rate.}
		\label{fig:MNIST_RD}
	\end{center}
	\vskip -0.25in
\end{figure*}

More concretely, we use an encoder $f$ and a decoder $g$, both parametrized by deep neural nets (DNNs). The encoder maps the input $x$ into a latent feature vector $f(x)$, whose entries are then uniformly quantized to $L$ levels
to obtain the representation $\hat{f}(x)$. The decoder outputs a reconstruction $\hat{x} = g(\hat{f}(x))$. To enable back-propagation through the quantizer, we use the differentiable relaxation of \cite{mentzer2018conditional}. Note that this relaxation affects only the gradient computation through the quantizer during back-propagation, but not the forward-pass ``hard'' quantization.

As in recent perceptual-quality driven lossy compression schemes \cite{tschannen2018deep,agustsson2018generative}, the rate is controlled by the dimension $dim$ of the encoder's output $f(x)$, and the number of levels $L$ used for quantizing each of its entries, such that $R \le dim\times\log_2(L)$. Note that this only upper-bounds the best achievable rate, as lossless compression of $\hat{f}(x)$ would potentially further reduce the representation's size. However, it significantly simplifies the scheme, and was found to be only slightly sub-optimal \cite{agustsson2018generative}.

For any fixed rate, we train the encoder-decoder to minimize a loss comprising a weighted combination of the expected distortion and the perception index,
\begin{equation}
\mathbb{E}[\Delta(X,g(\hat{f}(X)))] + \lambda d_{\text{W}}(p_X,p_{\hat{X}}).
\end{equation}
Here, $\hat{X} = g(\hat{f}(X))$, $d_{\text{W}}(\cdot,\cdot)$ is the Wasserstein distance, and $\lambda$ is a tuning parameter, which we use to control the balance between perception and distortion. The perceptual quality term can be optimized with the framework of generative adversarial networks (GAN) \cite{goodfellow2014generative} by introducing an additional (discriminator) DNN $h:\mathcal{X} \rightarrow \mathbb{R}$ and minimizing
\begin{align}\label{eq:objective}
\mathbb{E}[\Delta(X,g(\hat{f}(X)))]
+ \lambda \max_{h \in \mathcal{F}}\! \left(\!\mathbb{E}[h(X)] - \mathbb{E}[h(g(\hat{f}(X)))] \!\right)
\end{align}
where in our specific case of a Wasserstein GAN \cite{arjovsky2017wasserstein}, $\mathcal{F}$ denotes the class of bounded $1$-Lipschitz functions. As usual, all expectations are replaced by sample means, the constraint $h \in \mathcal{F}$ is replaced by a gradient penalty \cite{gulrajani2017improved}, and the loss is minimized by alternating between minimization w.r.t.~$f,g$ while holding $h$ fixed and maximization w.r.t.~$h$ while holding $f,g$ fixed.

\begin{figure*}[!t]
	\vskip 0.05in
	\centering
	\subfigure[]{%
		\includegraphics[width=.33\textwidth]{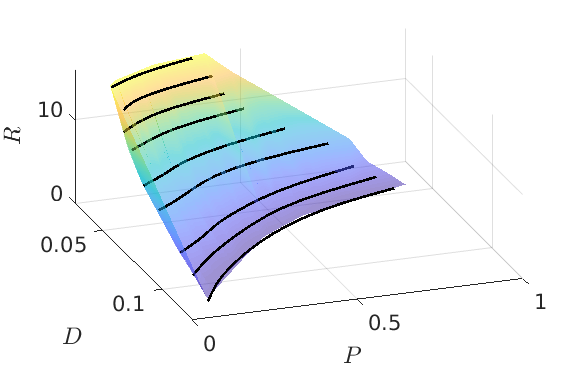} \label{fig:MNIST_1} 
	} 
	\subfigure[]{%
		\includegraphics[width=.31\textwidth]{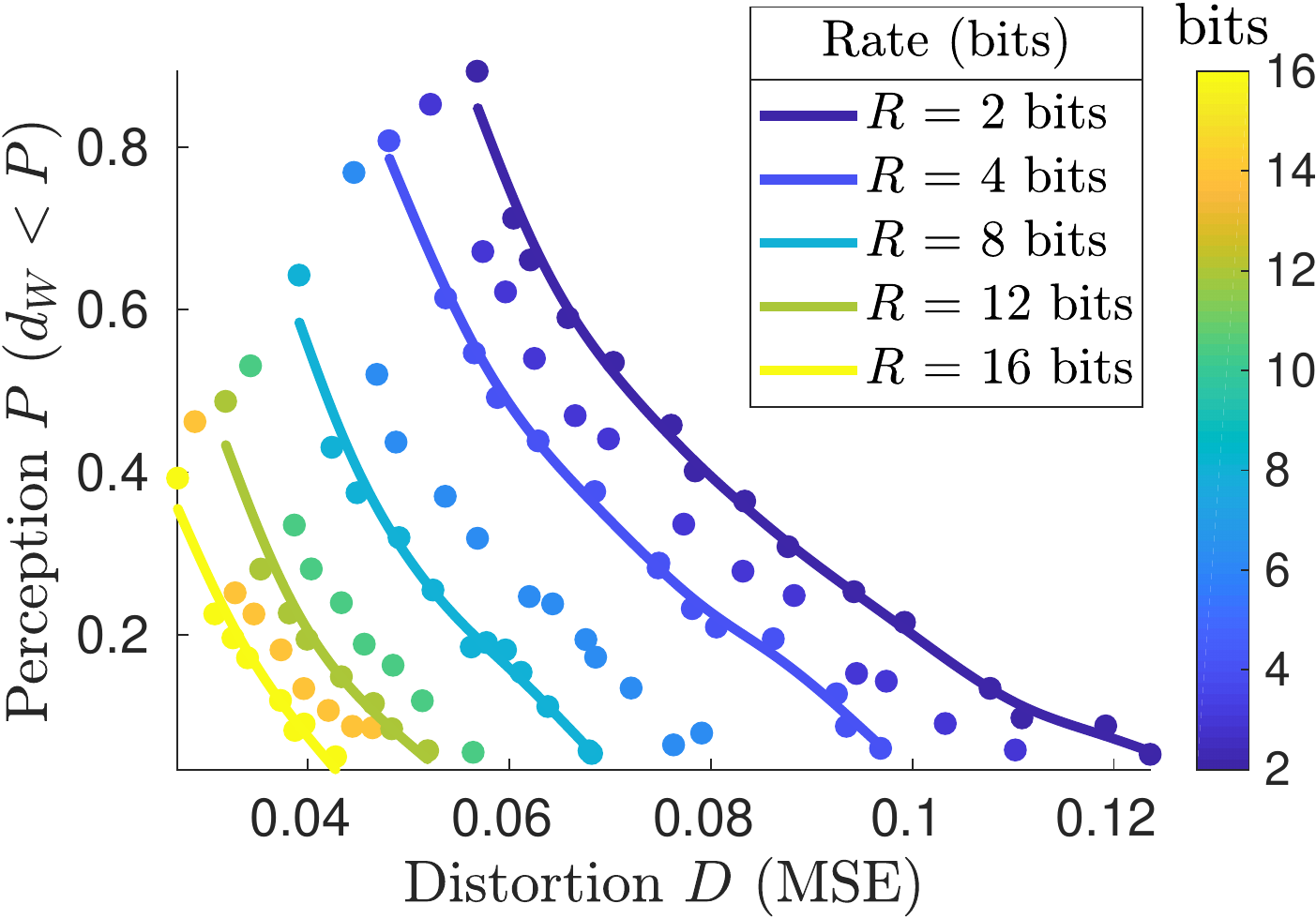}
		\label{fig:MNIST_2} 
	} 
	\hspace{0.2cm}
	\subfigure[]{%
		\includegraphics[width=.32\textwidth]{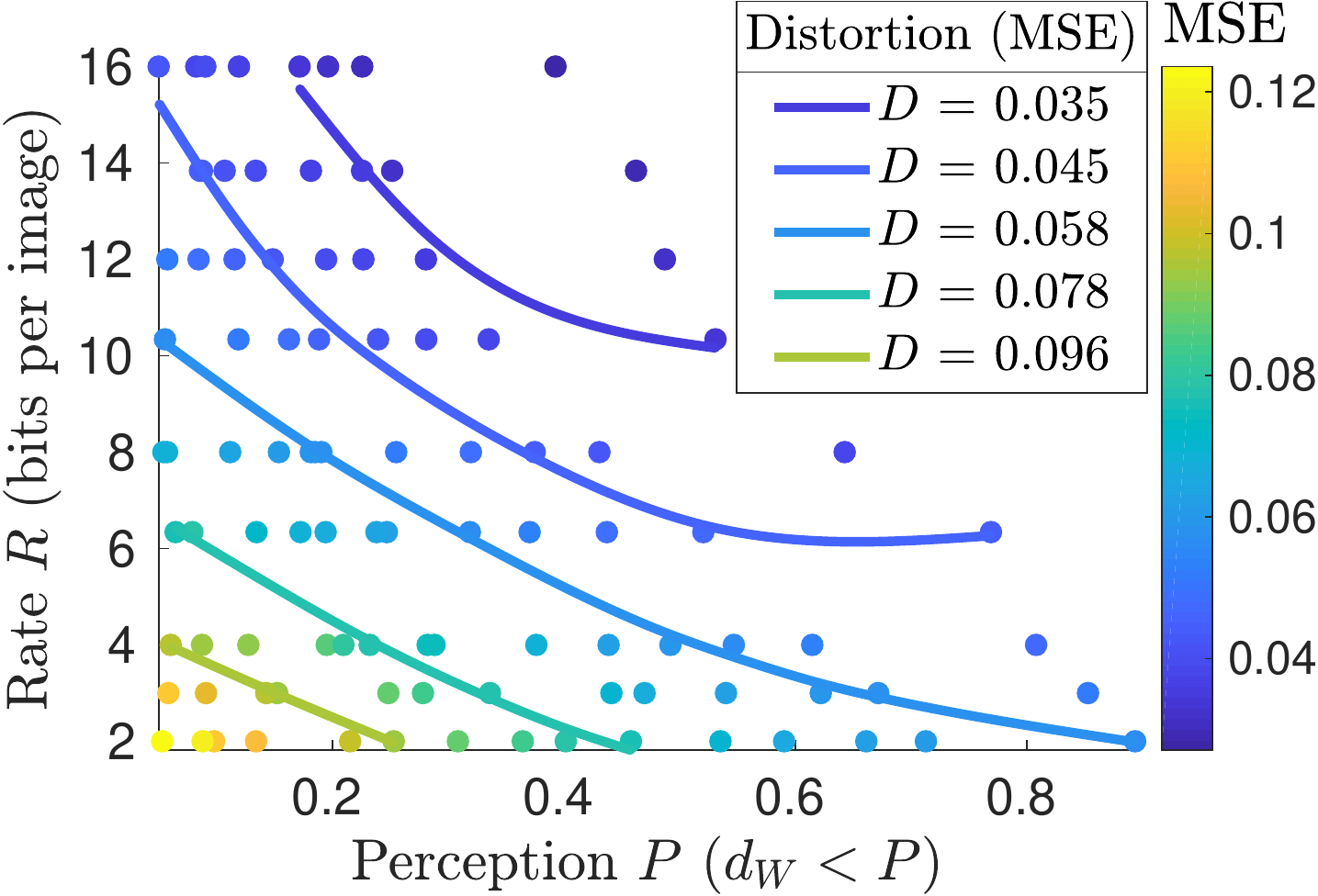}
		\label{fig:MNIST_3} 
	} 
	\vskip -0.05in
	\caption{\textbf{The rate-distortion-perception function of MNIST images.} (a) Equi-rate lines plotted on $R(D,P)$ highlight the tradeoff between distortion and perceptual quality at any constant rate. (b) Cross sections of $R(D,P)$ along perception-distortion planes show that this tradeoff becomes stronger at low bit-rates. (c) Cross-sections of $R(D,P)$ along rate-perception planes highlght that at any \emph{constant} distortion, the perceptual quality can be improved by increasing the rate.}
	\label{fig:MNIST_figs}
\end{figure*}

To achieve good perceptual quality, especially at low rates, it is essential that the decoder be stochastic \cite{tschannen2018deep}. This is commonly carried out by an additional random noise input. Yet, deep generative models in the conditional setting tend to ignore this type of stochasticity \cite{zhu2017unpaired,zhu2017toward,mathieu2016deep}. Tschannen et al.~\yrcite{tschannen2018deep} remedy this by applying a two-stage training scheme, which indeed promotes the use of stochasticity within the decoder, but can lead to sub-optimal results. Here, instead of concatenating a noise vector $n$ to the encoder's output $\hat{f}(x)$, we \emph{add} it, so that the decoder in fact operators on the noisy representation $\hat{f}(x)+n$. This \emph{does not} lead to loss of information, as the noise $n$ is drawn from a uniform distribution $U(-\tfrac{\alpha}{2},\tfrac{\alpha}{2})$, with $\alpha$ smaller than the quantization bin size. Thus, different coded representations $\hat{f}(x)$ do not ``mix-up'', and can always be distinguished from one another. This scheme urges the decoder to utilize the stochastic input, while allowing end-to-end training in a one-step manner.

\subsection{Squared-Error Distortion}\label{sec:exp_MNIST}

We begin by experimenting with the squared-error distortion $\Delta(x,\hat{x})=\|x-\hat{x}\|^2$. We train 98 encoder-decoder pairs on the MNIST handwritten digit dataset \cite{lecun1998gradient}, while varying the encoder's output dimension $dim$ and number of quantization levels $L$ to control the rate $R$, and the tuning coefficient $\lambda$ to achieve different balances between distortion and perceptual quality. A list of all combinations of $(dim, L, \lambda)$ used, along with all other training details can be found in the Supplementary Material.

The left side of Fig.~\ref{fig:MNIST_RD} plots the 98 trained encoder-decoder pairs on the rate-distortion plane, with the perceptual quality indicated by color coding and rate measured in bits per digit. The perceptual quality is quantified by the final discriminator loss\footnote{$\mathcal{L}_{\text{dis}} = \tfrac{2}{N} \left( \sum_{i=1}^{N/2}h(x_i) - \sum_{i=N/2+1}^{N}h(g(\hat{f}(x_i))) \right)$, where $\{x_i\}_{i=1}^N$ are the test samples.}, which approximates the Wasserstein distance $d_{\text{W}} (p_X,p_{\hat{X}})$. We plot an approximation of Shannon's rate-distortion function (obtained with $\lambda=0$), and two additional rate-distortion curves with (approximately) constant perceptual quality\footnote{We plot a smoothing spline calculated over the set of points which satisfy the constraint $d_{\text{W}}(p_X,p_{\hat{X}}) \le P$ and have the minimal distortion among all points with the same rate.}. As can be seen, the rate-distortion curve elevates when constraining the perceptual quality to be good. This demonstrates once again that we can improve the perceptual quality w.r.t.\@ that obtained on Shannon's rate-distortion curve, yet this must come at the cost of a higher rate and\slash or distortion. Notice that the perception index is not constant along Shannon's function; it increases (worse quality) towards lower bit-rates.

On the right side of Fig.~\ref{fig:MNIST_RD}, we depict the outputs of encoder-decoder pairs along Shannon's rate-distortion function, and along the two equi-perception curves shown on the left. It can be seen that as the rate decreases, the perceptual quality of the reconstructions along Shannon's function degrades. However, this is avoided when constraining the perceptual quality, which results in visually pleasing reconstructions even at extremely low bit-rates. Notice that this increased perceptual quality does not imply increased accuracy, as at low bit rates (e.g., $2$ bits), most reconstructions fail to preserve even the identity of the digit. Yet, while the encoder-decoder pairs on Shannon's rate-distortion curve are more accurate on average, no doubt that the perceptually-constrained encoder-decoder pairs are favorable in terms of perceptual quality. Also, notice that at a rate of $2$ bits, the outputs of the perceptually-constrained encoder-decoder pairs are all distinct, even though there are only $4$ code words (as can be seen for Shannon's encoder-decoder). This shows that the decoder effectively utilizes the noise.

Figure~\ref{fig:MNIST_figs} depicts the function $R(D,P)$ in 3-dimensions, as well as its cross sections along the other axis aligned planes. In Fig.~\ref{fig:MNIST_1}, the \emph{curved} equi-rate lines show the tradeoff between distortion and perceptual quality. This is also apparent in Fig.~\ref{fig:MNIST_2}, which shows cross sections along perception-distortion planes at different rates. As can be seen, the tradeoff becomes stronger at low bit-rates. Figure \ref{fig:MNIST_3} shows the counter-intuitive tradeoff between rate and perception. That is, at \emph{constant} distortion, the perceptual quality can be improved by increasing the rate.

\subsection{Advanced Distortion Measures}\label{sec:exp_perceptual}

\begin{figure}[t]
	\vskip 0.05in
	\begin{center}
		\centerline{\includegraphics[width=0.9\columnwidth]{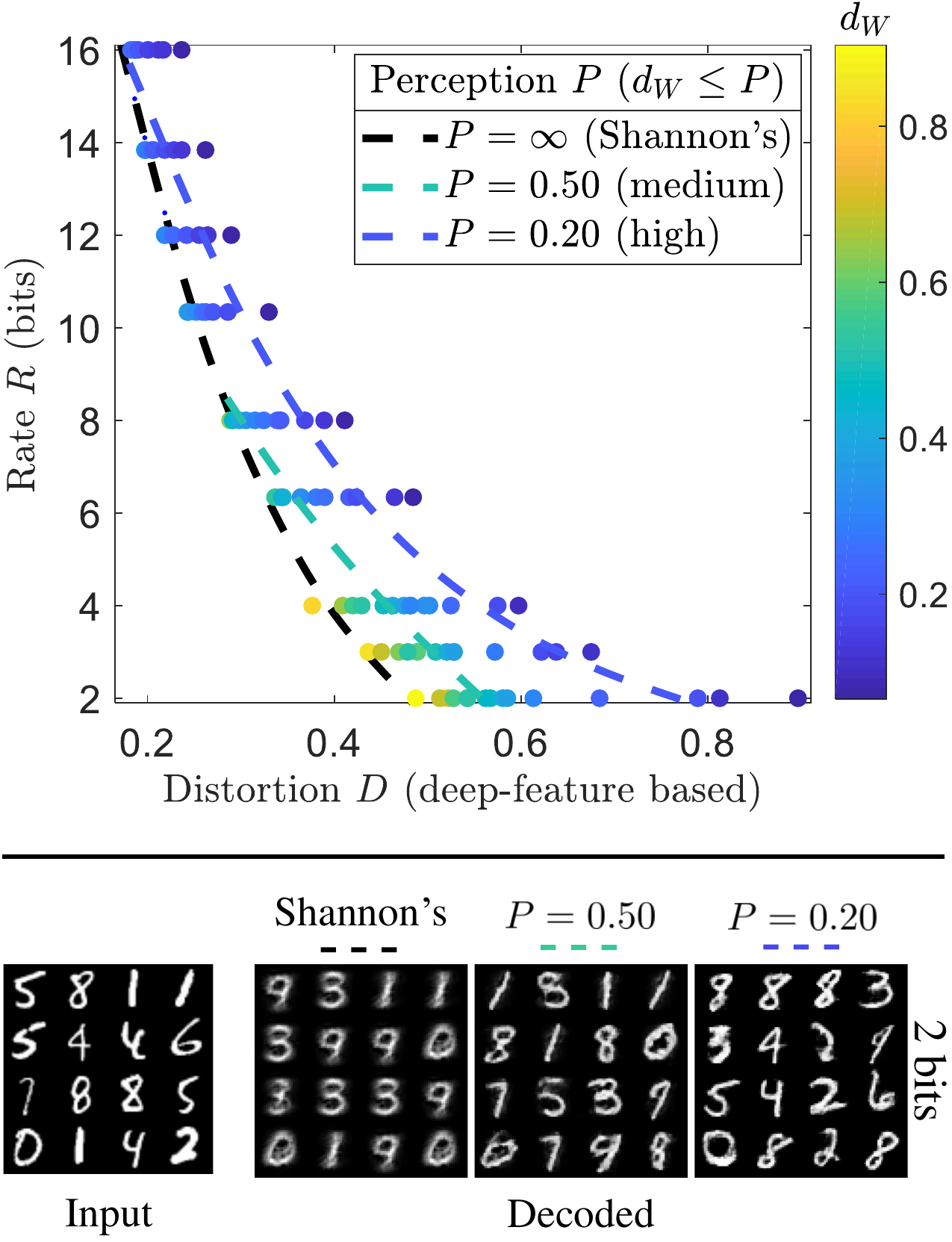}}
		\caption{\textbf{Replacing the MSE with the deep feature based distortion.} We repeat the experiment of Fig.~\ref{fig:MNIST_RD}, while replacing the squared-error distortion with the deep-feature based distortion of Johnson et al.~\yrcite{johnson2016perceptual}. \emph{Top}: The rate-distortion curves elevate when constraining the perceptual quality, demonstrating that the use of this advanced distortion measure \emph{does not} eliminate the tradeoff. \emph{Bottom}: Even with this popular advanced distortion, is it still beneficial to compromise distortion and constrain for improved perceptual quality. Nevertheless, the tradeoff does appear to be a bit weaker than in Fig.~\ref{fig:MNIST_RD}, as minimizing distortion alone (Shannon's curve) is now somewhat more pleasing visually.}
		\label{fig:MNIST_RD_perc}
	\end{center}
	\vskip -0.25in
\end{figure}

The peak-signal-to-noise ratio (PSNR), which is a rescaling of the MSE, is still the most common quality measure in image compression. Yet, it is well-known to be inadequate for quantifying distortion as perceived by humans \cite{wang2009mean}. Over the past decades, there has been a constant search for better distortion criteria, ranging from the simple SSIM\slash MS-SSIM \cite{wang2003multiscale,wang2004image} to the recently popular deep-feature based distortion \cite{johnson2016perceptual,zhang2018unreasonable}. Interestingly, the perceptual quality along Shannon's classical rate-distortion function is not perfect for nearly any distortion measure (see property 3 in Theorem \ref{thm:convexity}). This implies that perfect perceptual quality cannot be achieved by merely switching to more advanced distortion criteria, but rather requires directly optimizing the perception index (e.g.~using GAN-based schemes). 
This is not to say that the function $R(D,P)$ is the same for all distortion measures. The strength of the tradeoff can certainly decrease for distortion criteria which capture more semantic similarities \cite{blau2018perception}.

We demonstrate this by repeating the experiment of Sec.~\ref{sec:exp_MNIST}, while replacing the squared-error distortion by the deep-feature based distortion of~\cite{ledig2017photo}, i.e.,
\begin{equation}\label{eq:perc_loss}
\Delta(x,\hat{x}) = \|x-\hat{x}\|^2 + \alpha \| \Psi(x) - \Psi(\hat{x}) \|^2,
\end{equation}
where $\Psi(x)$ is the output of an intermediate DNN layer for input $x$. Here we take the second conv-layer output of a $4$-layer DNN, which we pre-trained to achieve over $99\%$ classification accuracy on the MNIST test set. 
All training details appear in the Supplementary Material.

Figure \ref{fig:MNIST_RD_perc} plots 98 encoder-decoder pairs on the rate-distortion plane, trained exactly as in Fig.~\ref{fig:MNIST_RD}, but this time with the loss~\eqref{eq:perc_loss} instead of MSE. As can be seen, here too the rate-distortion curves elevate when constraining the perceptual quality, demonstrating that the use of advanced distortion measures \emph{does not} eliminate the tradeoff. From the decoded outputs, however, it is evident that the tradeoff here is somewhat weaker, as minimizing distortion alone (Shannon's) appears a bit more visually pleasing compared to Fig.~\ref{fig:MNIST_RD} (though still with reduced variability and more blur than the perception constrained reconstruction).
\subsection{Related Work}
Our theoretical analysis and experimental validation help explain some of the observations reported in the recent literature. Specifically, a lot of research efforts have been devoted to optimizing the rate-distortion function \eqref{eq:RD} using deep nets \cite{toderici2016variable,toderici2017full,agustsson2017soft,balle2017end,minnen2018joint,li2018learning}. Some papers explicitly targeted high perceptual quality. One line of works did so by choosing the distortion criterion to be some advanced full-reference measure, like SSIM\slash MS-SSIM \cite{balle2018variational,mentzer2018conditional,johnston2018improved}, normalized Laplacian pyramid \cite{balle2016end} and deformation-aware sum of squared differences (DASSD)~\cite{rott2018deformation}. While beneficial, these methods could not demonstrate high perceptual quality at very low bit rates, which aligns with our theory. Another line of works incorporated generative models, which explicitly encourage the distribution of outputs to be similar to that of natural images (decreasing the divergence in \eqref{eq:perceptualQuality}). This was done on an image patch level \cite{rippel2017real}, on reduced-size (thumbnail) images \cite{tschannen2018deep,santurkar2018generative}, on a full-image scale \cite{agustsson2018generative}, and as a post-processing step \cite{galteri2017deep}. In particular, Tschannen et al.~\yrcite{tschannen2018deep} propose a practical method for distribution-preserving compression ($P=0$ in our terminology). These methods managed to obtain impressive perceptual quality at very low bit rates, but not without a substantial sacrifice in distortion, as predicted by our theory. Finally, we note that rate-distortion analysis (with a specific distortion) has also been used in the context of generative models \cite{alemi2018fixing}, which target $p_{\hat{X}}=p_X$ (i.e., $P=0$). Our results hold for arbitrary distortions and arbitrary $P$.
\section{Conclusion}
We proved that in lossy compression, perceptual quality is at odds with rate and distortion. Specifically, any attempt to keep the statistics of decoded signals similar to that of source signals, will result in a higher distortion or rate. We characterized the triple tradeoff between rate, distortion and perception, and empirically illustrated its manifestation in image compression. Our observations suggest that 
comparing methods based on their rate-distortion curves alone may be misleading. A more informative evaluation must also include some (no-reference) perceptual quality measure.

\section*{Acknowledgements}
This research was supported in part by the Israel Science Foundation
(grant no. 852/17) and by the Ollendorf Foundation.



\bibliographystyle{icml2019}

\clearpage
\onecolumn
\thispagestyle{empty}
\appendix

\begin{center}
\hrule height1pt \vskip .25in
{\Large\bf Rethinking Lossy Compression: The Rate-Distortion-Perception Tradeoff\\ Supplementary Material}
\vskip .22in \hrule height1pt \vskip .3in
\end{center}

\onehalfspacing

In this supplemental, we first provide proofs for Theorems~\ref{thm:convexity} and~\ref{thm:bound}. We then prove the relation between the rate of an encoder-decoder pair and the rate-distortion-perception function in \eqref{eq:RDP} for a memoryless stationary source. Next, we derive the rate-distortion-perception function $R(D,P)$ of a Bernoulli random variable (see Sec.~\ref{sec:Bernoulli}), which appears in \eqref{eq:Bern_RDP}. In the following section, we specify all training and architecture details for the experiments in Sec.~\ref{sec:exp}. Finally, we include details on the choice of the perceptual loss in the experiment of Sec.~\ref{sec:exp_perceptual}.

\section{Proof of Theorem~\ref{thm:convexity}}
The proof of this theorem follows closely that of its rate-distortion analogue (Cover \& Thomas \yrcite{cover2012elements}, 2nd ed., p.~316). 

\paragraph{Monotonicity}
The value $R(P,D)$ is the minimal mutual information $I(X,\hat{X})$ over a constraint set whose size increases with $D$ and $P$. This implies that the function $R(D,P)$ is non-increasing in $D$ and $P$.

\paragraph{Convexity}
Here, we assume that \textbf{A1} holds. That is, the divergence $d(p,q)$ in \eqref{eq:RDP} is convex in its second argument, so that for any $\lambda \in [0,1]$,
\begin{equation}
d(p,\lambda q_1 + (1-\lambda) q_2) \le \lambda d(p,q_1) + (1-\lambda) d(p,q_2).
\end{equation}
To prove the convexity of $R(D,P)$, we will show that
\begin{equation}\label{eq:proofObjective}
	\lambda R(D_{1}, P_1)+(1-\lambda)R(D_2,P_2) \ge  R(\lambda D_{1}+(1-\lambda)D_{2},\lambda P_{1}+(1-\lambda)P_{2}),
\end{equation}
for all $\lambda \in [0,1]$. First, by definition, the left hand side of~\eqref{eq:proofObjective} can be written as
\begin{equation}\label{eq:pr1}
	\lambda I(X, \hat{X}_{1}) + (1-\lambda)I(X, \hat{X}_{2}),
\end{equation}
where $\hat{X}_1$ and $\hat{X}_2$ are defined by
\begin{align}\label{eq:defX1}
	p_{\hat{X}_1 \vert X} &= \arg\min_{p_{\hat{X}|X}} \,\, I(X,\hat{X}) \quad \text{s.t.} \quad \mathbb{E}[\Delta(X,\hat{X})] \le D_1,\, d(p_X, p_{\hat{X}}) \le P_1,  \\
	p_{\hat{X}_2 \vert X} &= \arg\min_{p_{\hat{X}|X}} \,\, I(X,\hat{X}) \quad \text{s.t.} \quad \mathbb{E}[\Delta(X,\hat{X})] \le D_2,\, d(p_X, p_{\hat{X}}) \le P_2.
\end{align}
Since $I(X,\hat{X})$ is convex in $p_{\hat{X}|X}$ for a fixed $p_X$ (Cover \& Thomas \yrcite{cover2012elements}, 2nd ed., p.~33),
\begin{equation}\label{eq:pr2}
	\lambda I(X, \hat{X}_{1}) + (1-\lambda)I(X, \hat{X}_{2}) \ge I(X, \hat{X}_{\lambda}),
\end{equation}
where $\hat{X}_{\lambda}$ is defined by
\begin{equation}\label{eq:define_XhatLambda}
	p_{\hat{X}_{\lambda}\vert X}=\lambda p_{\hat{X}_1\vert X}+\left(1-\lambda \right)p_{\hat{X}_2\vert X}.
\end{equation}
Denoting $D_\lambda = \mathbb{E}[\Delta(X,\hat{X}_\lambda)]$ and $P_\lambda = d(p_X,p_{\hat{X}_\lambda})$, we have that
\begin{equation}\label{eq:pr3}
	I(X, \hat{X}_{\lambda}) \ge \min_{p_{\hat{X} \vert X}} \left\{ I(X,\hat{X}) \, : \, \mathbb{E}[\Delta(X,\hat{X})] \le D_\lambda, \, d(p_X,p_{\hat{X}_\lambda}) \le P_\lambda \right\}
	= R(D_{\lambda}, P_\lambda),
\end{equation}
because $\hat{X}_{\lambda}$ is in the constraint set. The divergence $d(p,q)$ is assumed to be convex in the second argument, thus
\begin{align}\label{eq:pr4a}
P_\lambda &= d(p_X,p_{\hat{X}_\lambda}) \nonumber\\
&\le \lambda d(p_X,p_{X_1}) + (1-\lambda) d(p_X,p_{X_2}) \nonumber\\
&\le \lambda P_1 + (1-\lambda) P_2.
\end{align}
Similarly,
\begin{align} \label{eq:pr5}
	D_\lambda &= \mathbb{E} \left[\Delta(X,\hat{X_\lambda})\right] \nonumber \\
	&\overset{\mathrm{(a)}}{=} \mathbb{E} \left[ \mathbb{E} \left[\Delta(X,\hat{X_\lambda}) \vert X \right]  \right] \nonumber \\
	&\overset{\mathrm{(b)}}{=} \mathbb{E} \left[ \lambda \mathbb{E} \left[\Delta(X,\hat{X_1}) \vert X \right] + (1-\lambda) \mathbb{E} \left[ \Delta(X,\hat{X_2}) \vert X \right]  \right] \nonumber \\
	&\overset{\mathrm{(c)}}{=} \lambda \mathbb{E} [\Delta(X,\hat{X}_1)] + \left(1-\lambda\right) \mathbb{E}[\Delta(X,\hat{X}_2)] \nonumber\\
	&\le \lambda D_1 + (1-\lambda)D_2,
\end{align}
where (a) and (c) are according to the law of total expectation, and (b) is by \eqref{eq:define_XhatLambda}. 
Therefore, since $R(D,P)$ is non-increasing in $D$ and $P$, we have from \eqref{eq:pr4a} and \eqref{eq:pr5} that
\begin{equation}\label{eq:pr4}
	R(D_{\lambda},P_\lambda) \ge R(\lambda D_{1}+(1-\lambda)D_2, \lambda P_{1}+(1-\lambda)P_2).
\end{equation}
Combining \eqref{eq:pr1}, \eqref{eq:pr2}, \eqref{eq:pr3} and \eqref{eq:pr4} proves \eqref{eq:proofObjective}, thus proving that $R(D,P)$ is convex.

\paragraph{Dependence on the perceptual quality}
Here, we assume that \textbf{A2} holds. In particular, this implies that the function $k(z) = \mathbb{E}_{X \sim p_X}[\Delta(X,z)]$ does not attain its minimum over the entire support of $p_X$. 
To prove that $R(\cdot,0) \ne R(\cdot,\infty)$, let us assume to the contrary that $R(\cdot,0) = R(\cdot,\infty)$. This implies that for any distortion level, minimizing the rate without a constraint on perception ($P=\infty$), leads to perfect perceptual quality, $p_{\hat{X}}=p_X$, just like with a perfect perception constraint $P=0$. Let us examine the solution specifically at the distortion level $D^*$ defined by
\begin{equation}
D^* = \min_{p_{\hat{X}|X}} \,\, \mathbb{E}_{(X,\hat{X}) \sim p_{X,\hat{X}}}[\Delta(X,\hat{X})] \quad \text{s.t.} \quad I(X,\hat{X})=0.
\end{equation}
Notice that since $I(X,\hat{X})=0$ in this case, $X$ and $\hat{X}$ are independent, so that $p_{\hat{X}|X}=p_{\hat{X}}$. Therefore
\begin{align}\label{eq:min_dist_R0}
D^* &= \min_{p_{\hat{X}}} \,\, \mathbb{E}_{(X,\hat{X}) \sim p_{X}p_{\hat{X}}}[\Delta(X,\hat{X})] \nonumber\\
&= \min_{p_{\hat{X}}} \,\, \mathbb{E}_{\hat{X} \sim p_{\hat{X}}}[\mathbb{E}_{X \sim p_X}[\Delta(X,\hat{X})]] \nonumber\\
&= \min_{p_{\hat{X}}} \,\, \mathbb{E}_{\hat{X} \sim p_{\hat{X}}}[k(\hat{X})].
\end{align}
Clearly, the $p_{\hat{X}}$ which minimizes \eqref{eq:min_dist_R0} cannot assign positive probability outside the set where $k(z)$ attains its minimal value. Namely, the support of $p_{\hat{X}}$ must be contained in the set $S$ defined by
\begin{equation}
S = \{z \in \arg\min_{\tilde{z}} \,\, k(\tilde{z}) \}.
\end{equation}
But since our encoder-decoder pair achieves perfect perceptual quality, i.e.~$p_{\hat{X}}=p_X$, this implies that $\text{support}\{p_X\} \subset S$, contradicting Assumption \textbf{A2}.

\section{Proof of Theorem~\ref{thm:bound}}
Assume the MSE distortion, and consider any $(R,D)$ pair on Shannon's classic rate-distortion function (corresponding to $R(D,\infty)$ on the rate-distortion-perception function), and the encoder-decoder mapping $p_{\hat{X}|X}$ which achieves this $(R,D)$ pair. We will prove the theorem by explicitly constructing a modified encoder-decoder, which achieves perfect perceptual quality and has only twice the distortion. To do this, we concatenate a post-processing mapping $p_{\tilde{X}|\hat{X}}$ to produce a new decoded output $\tilde{X}$ by drawing from the posterior distribution $p_{X|\hat{X}}$. That is,
\begin{equation}
p_{\tilde{X}|\hat{X}}(\tilde{x}|\hat{x}) = p_{X|\hat{X}}(\tilde{x}|\hat{x}) =  \frac{p_{\hat{X}|X}(\hat{x}|\tilde{x})p_{X}(\tilde{x})}{p_{\hat{X}}(\hat{x})}.
\end{equation}
The distribution $p_{\tilde{X}}$ of this new output $\tilde{X}$ is identical to the distribution of the source signal $p_X$, as
\begin{equation}
p_{\tilde{X}}(z) = \int p_{\tilde{X}|\hat{X}}(z|\hat{x}) p_{\hat{X}}(\hat{x}) d\hat{x} = \int p_{X|\hat{X}}(z|\hat{x}) p_{\hat{X}}(\hat{x}) d\hat{x} = p_X(z).
\end{equation}
Therefore, $d(p_X,p_{\tilde{X}})=0$, showing that it achieves perfect perceptual quality. The MSE distortion of the modified encoder-decoder is given by
\begin{align}\label{eq:D_tilde}
\tilde{D} &= \mathbb{E}[\| X-\tilde{X} \|^2] \nonumber \\
&= \mathbb{E}[\|X\|^2] -2 \mathbb{E}[ X^T\tilde{X}] + \mathbb{E}[\|\tilde{X}\|^2] \nonumber \\
&\overset{\mathrm{(a)}}{=} \mathbb{E}[\|X\|^2] -2 \mathbb{E}[\mathbb{E}[ X^T\tilde{X}|\hat{X}]] + \mathbb{E}[\mathbb{E}[\|\tilde{X}\|^2|\hat{X}]] \nonumber \\
& \overset{\mathrm{(b)}}{=} \mathbb{E}[\|X\|^2] -2 \mathbb{E}[\|\mathbb{E}[ X|\hat{X}]\|^2] + \mathbb{E}[\mathbb{E}[\|X\|^2|\hat{X}]] \nonumber \\
&= \mathbb{E}[\|X\|^2] - 2\mathbb{E}[ \|\mathbb{E}[X|\hat{X}]\|^2]) + \mathbb{E}[\|X\|^2|]\nonumber \\
&= 2(\mathbb{E}[\|X\|^2] - \mathbb{E}[\| \mathbb{E}[X|\hat{X}]\|^2]) \nonumber \\
&= 2\,\mathbb{E}[\| X - \mathbb{E}[X|\hat{X}] \|^2] \nonumber \\
& \overset{\mathrm{(c)}}{=}  2\,\mathbb{E}[\| X - \hat{X} \|^2] = 2D,
\end{align}
where in (a) we used the law of total expectation, in (b) we used the fact that $X$ and $\tilde{X}$ are independent given $\hat{X}$ and both have distribution $p_X$, and in (c) we used that fact that $\hat{X}=\mathbb{E}[X|\hat{X}]$ as otherwise it would not have lied on the rate-distortion curve (replacing $\hat{X}$ by $\mathbb{E}[X|\hat{X}]$ would lead to a lower MSE without increasing the rate).

The mutual information between the source $X$ and the modified decoded signal $\tilde{X}$ satisfies
\begin{align}\label{eq:I_X_tilde}
I(X;\tilde{X}) \le I(X;\hat{X}),
\end{align}
due to the data processing inequality for the Markov chain $X\rightarrow \hat{X} \rightarrow \tilde{X}$.

Putting it together, we get
\begin{align}
R(D,\infty) &= \min_{p_{\hat{X}|X}} \, \{ I(X,\hat{X}) \, : \, \mathbb{E}[\Delta(X,\hat{X})] \le D \} \nonumber \\
&\overset{\mathrm{(f)}}{\ge} I(X,\tilde{X}) \nonumber \\
&\overset{\mathrm{(g)}}{\ge} \min_{p_{\hat{X}|X}} \, \{ I(X,\hat{X}) \, : \, \mathbb{E}[\Delta(X,\hat{X})] \le \tilde{D}, d(p_X, p_{\hat{X}}) \le 0 \} \nonumber \\
&= R(\tilde{D},0) \nonumber \\
&\overset{\mathrm{(h)}}{\ge} R(2D,0),
\end{align}
where (f) is due to \eqref{eq:I_X_tilde}, (g) is since $p_{\tilde{X}|X}$ is in the constraint set, and (h) is justified by \eqref{eq:D_tilde} and the fact that $R(D,P)$ is non-increasing in $D$ (see Theorem \ref{thm:convexity}). This proves that $R(D,0) \le R(\tfrac{1}{2}D,\infty)$.

\section{Perception aware lossy compression of a memoryless stationary source}
We now prove that when compressing a memoryless stationary source with average distortion $D$ and average perception index $P$, the rate is lower bounded by $R(D,P)$. This proof follows closely that of its rate-distortion analogue (Cover \& Thomas \yrcite{cover2012elements}, 2nd ed., p.~316). 

Assume a memoryless stationary source. Given a source sequence $X^n$ comprising i.i.d.~variables $X_1,\ldots,X_n$ with distribution $p_X$, the encoder $f_n$ constructs an encoded representation with rate $R$ as $f_n : \mathcal{X}^n \rightarrow\{1,2,\ldots,2^{nR}\}$. The decoder $g_n$ outputs an estimate $\hat{X}^n$ of $X^n$ as $g_n:\{1,2,\ldots,2^{nR}\} \rightarrow \hat{\mathcal{X}}^n$. We are interested in the the average distortion of the reconstructions, $\frac{1}{n} \sum_{i=1}^n \Delta(X_i,\hat{X}_i)$, and in their average perceptual quality, $\frac{1}{n} \sum_{i=1}^n d(p_{X_i},p_{\hat{X}_i})$. Assume that
\begin{equation}\label{eq:D_P_levels}
\frac{1}{n} \sum_{i=1}^n \Delta(X_i,\hat{X}_i) \le D, \quad \frac{1}{n} \sum_{i=1}^n d(p_{X_i},p_{\hat{X}_i}) \le P.
\end{equation}
Then
\begingroup
\allowdisplaybreaks
\begin{align}
nR &\overset{\mathrm{(a)}}{\ge} H(f_n(X^n)) \nonumber \\
&\overset{\mathrm{(b)}}{\ge} H(f_n(X^n)) - H(f_n(X^n)|X^n) \nonumber \\
&= I(X^n;f_n(X^n)) \nonumber \\
&\overset{\mathrm{(c)}}{\ge} I(X^n,\hat{X}^n) \nonumber \\
&= H(X^n) - H(X^n|\hat{X}^n) \nonumber \\
&\overset{\mathrm{(d)}}{=} \sum_{i=1}^n H(X_i) - H(X^n|\hat{X}^n) \nonumber \\
&\overset{\mathrm{(e)}}{=} \sum_{i=1}^n H(X_i) - \sum_{i=1}^n H(X_i|\hat{X}^n,X_{i-1},\ldots,X_1) \nonumber \\
&\overset{\mathrm{(f)}}{\ge} \sum_{i=1}^n H(X_i) - \sum_{i=1}^n H(X_i|\hat{X}_i) \nonumber \\
&= \sum_{i=1}^n I(X_i,\hat{X}_i) \nonumber \\
&\overset{\mathrm{(g)}}{\ge} \sum_{i=1}^n R\left(\mathbb{E}[\Delta(X_i,\hat{X}_i)],d(p_{X_i},p_{\hat{X}_i})\right) \nonumber \\
&= n \left( \frac{1}{n} \sum_{i=1}^n R\left(\mathbb{E}[\Delta(X_i,\hat{X}_i)],d(p_{X_i},p_{\hat{X}_i})\right) \right) \nonumber \\
&\overset{\mathrm{(h)}}{\ge} n \, R \left( \frac{1}{n} \sum_{i=1}^n \mathbb{E}[\Delta(X_i,\hat{X}_i)], \frac{1}{n} \sum_{i=1}^n d(p_{X_i},p_{\hat{X}_i})\right) \nonumber \\
&\overset{\mathrm{(i)}}{\ge} n\, R(D,P),
\end{align}%
\endgroup
where (a) is since the size of the range of $f_n$ is $2^{nR}$, (b) is since $H(f_n(X^n)|X^n)>0$, (c) is from the data-processing inequality, (d) is since $X_i$ are independent, (e) is from the chain rule of entropy, (f) is since conditioning reduces entropy, (g) is from the definition of $R(D,P)$ in \eqref{eq:RDP}, (h) is from the convexity of $R(D,P)$ (see Theorem \ref{thm:convexity}) and Jensen's inequality, and (i) is from \eqref{eq:D_P_levels} and the fact that $R(D,P)$ is non-increasing in $D,P$ (see Theorem \ref{thm:convexity}). This proves that the rate of \emph{any} encoder-decoder pair having average distortion $\frac{1}{n} \sum_{i=1}^n \Delta(X_i,\hat{X}_i) = D$ and average perceptual quality $\frac{1}{n} \sum_{i=1}^n d(p_{X_i},p_{\hat{X}_i})=P$, is lower-bounded by $R(D,P)$, the rate-distortion-perception function evaluated at $D,P$.

To prove that the rate-distortion-perception function describes the optimal rate at distortion level $D$ and perceptual quality $P$, we would also have to prove that $R(D,P)$ is achievable, which we leave for future work. Yet, the proof that $R(D,P)$ lower-bounds the rate is sufficient for concluding that a tradeoff between rate, distortion and perception necessarily exists. Specifically, in Theorem \ref{thm:convexity} we prove that (subject to assumptions) the rate-distortion curve elevates when constraining for perceptual quality, i.e.~$R(\cdot,0) > R(\cdot,\infty)$. Now, Shannon's rate-distortion curve $R(\cdot,\infty)$ is known to be achievable (Cover \& Thomas \yrcite{cover2012elements}, 2nd ed., p.~318) and thus describes the optimal rate $R_S$ when \emph{not} constraining the perceptual quality. As shown above, $R(\cdot,0)$ lower-bounds the rate $R_P$ when constraining for perfect perceptual quality. Combining these, we get that $R_P>R_S$, indicating that constraining for perceptual quality necessarily leads to an increase in rate (for constant distortion level), thus illustrating the rate-distortion-perception tradeoff.

\section{Derivation of the rate-distortion-perception function $R(D,P)$ of a Bernoulli source}
Assume that $X \sim \text{Bern}(p)$ with $p \le \tfrac{1}{2}$. We seek a conditional distribution $p_{\hat{X}|X}$, which we parameterize by $a, b$ as
\begin{align}
&P(\hat{X}=0| X=0) = a,\\
&P(\hat{X}=0| X=1)= b,
\end{align}
that solves the rate-distortion-perception problem
\begin{equation}\label{eq:Bern_RDP_supp}
R(D,P) = \min_{a,b} \, I(X,\hat{X}) \quad \text{s.t.} \quad \mathbb{E}[\Delta(X,\hat{X})] \le D,\,\, d(p_X,p_{\hat{X}}) \le P.
\end{equation}
Here we concentrate on the case where $\Delta(\cdot,\cdot)$ is the Hamming distance, and $d(\cdot,\cdot)$ is the total-variation (TV) divergence.
The mutual information term $I(X,\hat{X})$ is given by
\begin{align}
I(X,\hat{X}) =& \sum_{x,\hat{x}\in\{0,1\}} P(X=x,\hat{X}=\hat{x}) \log \left( \frac{P(X=x,\hat{X}=\hat{x})}{P(X=x)P(\hat{X}=\hat{x})} \right) \nonumber \\
=&- a(1-p)\log \left( (1-p)+\frac{b}{a}p \right) - (1-a)(1-p)\log \left( (1-p)+\frac{1-b}{1-a}p \right) \nonumber\\
&- bp\log \left( \frac{a}{b}(1-p)+p \right) - (1-b)p\log \left( \frac{1-a}{1-b}(1-p)+p \right), \label{eq:I}
\end{align}
the Hamming distance term is given by
\begin{align}\label{eq:hamming}
d_{\text{H}}(X,\hat{X}) &= P(X=0,\hat{X}=1) + P(X=1,\hat{X}=0)
= (1-a) (1-p) + b p,
\end{align}
and the TV divergence term is given by
\begin{align}
d_{\text{TV}}(p_X,p_{\hat{X}}) &= \tfrac{1}{2} \sum_{z \in {0,1}} |p_X(z) - p_{\hat{X}}(z) | \nonumber\\
&= \tfrac{1}{2} (| P(X=0) - P(\hat{X}=0) | + | P(X=1) - P(\hat{X} = 1) |) \nonumber \\
&= |(1-a)(1-p) - bp |. \label{eq:TV_div}
\end{align}

\paragraph{Solution for $P=\infty$ (Shannon's rate-distortion problem)}
The function $R(D,\infty)$ for Shannon's classic rate-distortion problem is given by (see \cite{cover2012elements}, 2nd ed., p. 308)
\begin{equation}\label{eq:Shannon_sol}
R(D,\infty) =
\begin{cases}
H_b(p) - H_b(D) \quad & 0 \le D \le p,\\
0 & D>p,
\end{cases}
\end{equation}
where $H_b$ denotes the binary entropy $H_b(z) = -z\log(z) - (1-z) \log(1-z)$. This optimal solution is obtained by setting the parameters $a, b$ to 
\begin{equation}\label{eq:R-D_solution}
a_{S}(D) = 
\begin{cases}
\frac{(1-D)(1-p-D)}{(1-p)(1-2D)} \quad &D \le p,\\
1 & D > p,
\end{cases}
\quad \quad b_{S}(D) = 
\begin{cases}
\frac{D(1-p-D)}{p(1-2D)}  \quad &D \le p,\\
1 & D > p.
\end{cases}
\end{equation}

\paragraph{Solution for finite $P$ and $I(X,\hat{X})>0$}
We now move on to incorporate the additional perception constraint $d(p_X,p_{\hat{X}}) \le P$. First, notice that the distortion constraint $\mathbb{E}[\Delta(X,\hat{X})] \le D$ is always active when $I(X,\hat{X})>0$, since $I(X,\hat{X})=0$ is achievable for any $P$ when $D$ is not an active constraint\footnote{We can always set $p_{\hat{X}|X}(\hat{X}=\hat{x}|X=x)=p_X(\hat{x})$ (i.e.~a random draw from $p_X$ disregarding the given input $x$), which satisfies $d_{\text{TV}}(X,\hat{X})=0$ and leads to $I(X,\hat{X})=0$ since $X$ and $\hat{X}$ are independent in this case. Only a constraint on the distortion can prevent this solution from being viable.}. From \eqref{eq:hamming}, the fact that $d_{\text{H}}(X,\hat{X}) = D$ implies that
\begin{align}\label{eq:b_constraint}
b = \frac{D- (1 - a) (1-p)}{p}.
\end{align}
Substituting \eqref{eq:b_constraint} into \eqref{eq:TV_div}, we get that
\begin{equation}
d_{\text{TV}}(p_X,p_{\hat{X}}) = |2(1-a)(1-p) - D |. \label{eq:d_TV}
\end{equation}
Therefore, the constraint $d_{\text{TV}}(p_X,p_{\hat{X}}) \le P$ is satisfied when
\begin{equation}\label{eq:P_constraint_on_a}
-P \le 2(1-a)(1-p) - D \le P \quad \Rightarrow \quad 1-\frac{D+P}{2 (1-p)} \le a \le 1-\frac{D-P}{2 (1-p)}.
\end{equation}
Below, we show that the lower constraint of \eqref{eq:P_constraint_on_a} is never active (see \textbf{J1}).
The upper constraint is obviously active only when $a_S(D)$ of \eqref{eq:R-D_solution} does not satisfy the upper bound in \eqref{eq:P_constraint_on_a}, which happens when
\begin{equation}\label{eq:upper_constraint_a}
1-\frac{D-P}{2 (1-p)} < \frac{(1-D)(1-p-D)}{(1-p)(1-2D)} \quad \Rightarrow \quad D > \frac{P}{1+2P-2p} \triangleq D_1.
\end{equation}
Therefore, when $D \le D_1$ the solution is independent of $P$ and is given by \eqref{eq:Shannon_sol}. When $D > D_1$, the constraint $d_{\text{TV}}(p_X,p_{\hat{X}}) \le P$ is active, the upper constraint of \eqref{eq:P_constraint_on_a} is active, and thus
\begin{equation}\label{eq:a_sol}
a = 1-\frac{D-P}{2 (1-p)},
\end{equation}
and by substituting into \eqref{eq:b_constraint} we also get
\begin{equation}\label{eq:b_sol}
b = \frac{D+P}{2p}.
\end{equation}
Note that in \eqref{eq:upper_constraint_a} we assumed $D \le \tfrac{1}{2}$, below we will justify that this is always the case in this region (see \textbf{J2}).

Now, substituting $a,b$ from \eqref{eq:a_sol}, \eqref{eq:b_sol} back into \eqref{eq:I} we get
\begin{align}
I(X,\hat{X}) =& (1-p-\frac{D-P}{2})\log \left( \frac{1-p-\frac{D-P}{2}}{(1-p)(1-p+P)} \right) + (\frac{D-P}{2})\log \left( \frac{\frac{D-P}{2}}{(1-p)(p-P)} \right) \nonumber\\
&+ (\frac{D+P}{2})\log \left( \frac{\frac{D+P}{2}}{p(1-p+P)} \right) + (p-\frac{D+P}{2})\log \left( \frac{p-\frac{D+P}{2}}{p(p-P)} \right) \nonumber \\
=& (q-\alpha)\log \left( \frac{q-\alpha}{q(q+P)} \right) + \alpha\log \left( \frac{\alpha}{q(p-P)} \right) \nonumber \\
&+ \beta\log \left( \frac{\beta}{p(q+P)} \right) + (p-\beta)\log \left( \frac{p-\beta}{p(p-P)} \right)
\end{align}
where $q=1-p, \, \alpha = \tfrac{D-P}{2}$ and $\beta = \tfrac{D+P}{2}$. This can be further simplified to obtain
\begin{equation}
I(X,\hat{X}) = 2H_b(p) + H_b(p-P) - H_t(\alpha,p) - H_t(\beta,q),
\end{equation}
where $H_t(p_1,p_2)$ is the entropy of a ternary random variable (taking values in a three element alphabet) with probabilities $p_1, p_2, 1-p_1-p_2$.

\paragraph{Solution for finite $P$ and $I(X,\hat{X})=0$}
The function $R(D,P)$ is non-increasing in $D$ (see Theorem \ref{thm:convexity}), and will reach  $R(D,P)=0$ for $a=b$ since in this case $\hat{X}$ and $X$ are independent. From \eqref{eq:a_sol} and \eqref{eq:b_sol}, this happens when
\begin{equation}\label{eq:def_D0}
1-\frac{D-P}{2 (1-p)} = \frac{D+P}{2p} \quad \Rightarrow \quad D = 2p(1-p)+(2p-1)P = 2pq + (p-q)P \triangleq D_2,
\end{equation}
where for $D=D_2$ we get
\begin{equation}
a = b = (1-p)+P.
\end{equation}
From this point onward, the solution is fixed, as mutual information $I(X,\hat{X})$ is non-negative and we cannot further decrease the objective of \eqref{eq:Bern_RDP_supp}.

\paragraph{Overall solution}
Putting all the pieces together, the overall solution for $P < p$ is
\begin{align}
R(D,P) = 
\begin{cases}
H_b(p) - H_b(D) & D \le D_1  \\
2H_b(p) + H_b(p-P) - H_t(\tfrac{D-P}{2},p) - H_t(\tfrac{D+P}{2},q) \quad & D_1 < D \le D_2 \\
0 & D_2 < D
\end{cases}
\end{align}
where $D_1$ and $D_2$ are defined in \eqref{eq:upper_constraint_a} and \eqref{eq:def_D0}, respectively. For $P \ge p$, the solution is independent of $P$ and is given by the solution to Shannon's classic rate-distortion curve for a Bernoulli source in \eqref{eq:Shannon_sol} (see justification in \textbf{J3} below).

\subsection*{Additional justifications}
\paragraph{J1} 
The solution $a_S(D)$ in \eqref{eq:R-D_solution} does not satisfy this lower constraint of \eqref{eq:P_constraint_on_a} when
\begin{equation}
1-\frac{D+P}{2 (1-p)} > \frac{(1-D)(1-p-D)}{(1-p)(1-2D)}.
\end{equation}
When $P<\frac{1-2p}{2}$ this happens for $D < \frac{P}{2p+2P-1} < 0$, which never occurs as $D \in [0,1]$. When $P \ge \frac{1-2p}{2}$ this happens for $D > \frac{P}{2p+2P-1}$. However, since $\frac{P}{2p+2P-1} > D_1 = \frac{P}{1+2P-2p}$ for all $p \le \frac{1}{2}$ (which is our assumption), the upper constraint of \eqref{eq:P_constraint_on_a} will always become active before the lower constraint.

\paragraph{J2} 
Taking the derivative of $D_2 = 2p(1-p)+(2p-1)P$ with respect to $p$ we obtain
\begin{equation}
    \frac{\partial D_2}{\partial p} = 2 - 4p + 2P 
\end{equation}
which is non-negative since $p \le \tfrac{1}{2}$. Thus, $D_2$ is increasing in $p$ for all $P>0$, and its largest value in the range $p\in[0,\tfrac{1}{2}]$, which is $D_2 = \tfrac{1}{2}$, is obtained at $p=\tfrac{1}{2}$. Thus, in the region where $D_1 < D \le D_2$, it is ensured that $D \le \tfrac{1}{2}$.

\paragraph{J3} 
Taking the derivative of $D_1 = \frac{P}{1+2P-2p}$ with respect to $P$ we obtain\begin{equation}
    \frac{\partial D_1}{\partial P} = \frac{1-2p}{(1+2P-2p)^2}
\end{equation}
which is non-negative for $p\le \tfrac{1}{2}$, thus $D_1$ is non-decreasing in $P$. It is easy to see from \eqref{eq:def_D0} that $D_2$ in non-increasing in $P$ (for $p\le \tfrac{1}{2}$). Thus, $D_1(P)=D_2(P)$ for a single $P$, which is $P=p$. For any $P \ge p$, there are no $D$ satisfying $D_1 < D \le D_2$.

\section{Architecture and training parameters for the experiments in Sec.~\ref{sec:exp}}
The architecture of the encoder, decoder and discriminator nets used for compressing (and decompressing) the MNIST images in Sec.~\ref{sec:exp} is detailed in Table \ref{tab:architecture}. The optimization objective is given in \eqref{eq:objective}, where $\Delta(x,\hat{x})$ is the squared-error distortion in Sec.~\ref{sec:exp_MNIST}, and a combination of the squared-error and the ``perceptual loss'' of Johnson et al.~\yrcite{johnson2016perceptual} in Sec.~\ref{sec:exp_perceptual}. The encoder output dimension $dim$, the number of quantization levels $L$, and values of the tradeoff coefficient $\lambda$ in \eqref{eq:objective} used for training the $98$ encoder-decoder pairs appear in Table \ref{tab:comp_parameters}. The distortion term in \eqref{eq:objective} was also multiplied by a constant factor of $10^{-3}$ for the MSE term (in Sec.~\ref{sec:exp_MNIST} and Sec.~\ref{sec:exp_perceptual}) and factor of $5 \times 10^{-5}$ for the perceptual loss (in Sec.~\ref{sec:exp_perceptual}). For each $dim$ and $L$, an encodoer-decoder with $\lambda=0$ (only distortion, no adversarial loss) was trained for 25 epochs. The other encoder-decoder pairs with $\lambda>0$ continued training from this point for another 25 epochs. The ADAM optimizer was used with $\beta_1=0.5, \beta_2=0.9$. Batch size was $64$. Initial learning rates were $10^{-2}\slash 2\times10^{-4}$ for the encoder-decoder\slash discriminator updates in Sec.~\ref{sec:exp_MNIST}, and $5\times 10^{-3}\slash 2\times10^{-4}$ for the encoder-decoder\slash discriminator  updates in Sec.~\ref{sec:exp_perceptual}. These learning rates decreased by $\tfrac{1}{5}$ after 20 epochs. The convolutional\slash transposed-convolutional layers filter size (in the decoder and discriminator) was always $5$, except for the last convolutional layer in the decoder where the filter size was $4$. No padding was used in the decoder, and a padding of $2$ was used in each convolutional layer of the discriminator.

\begin{table}
	\centering
	\caption{Encoder, decoder, and discriminator architectures. FC is a fully-connected layer, Conv\slash ConvT is a convolutional\slash transposed-convolutional layer with ``st'' denoting stride,  BN is a batch-norm layer, and l-ReLU is a leaky-ReLU activation.}\label{tab:architecture}
	\small
	\begin{tabular}[t]{|c|c|}
		\hline
		\multicolumn{2}{c}{Encoder} \\
		\hline
		Size & Layer\\
		\hline \hline
		$28 \times 28 \times 1$ & Input  \\
		\hline
		$784$ & Flatten  \\
		\hline
		$512$ & FC, BN, l-ReLU\\
		\hline
		$256$ & FC, BN, l-ReLU\\
		\hline
		$128$ & FC, BN, l-ReLU\\
		\hline
		$128$ & FC, BN, l-ReLU\\
		\hline
		$dim$ & FC, BN, Tanh\\
		\hline
		$dim$ & Quantize\\
		\hline
	\end{tabular}
	\quad
	\begin{tabular}[t]{|c|c|}
		\hline
		\multicolumn{2}{c}{Decoder} \\
		\hline
		Size & Layer\\
		\hline \hline
		$dim$ & Input  \\
		\hline
		$128$ & FC, BN, l-ReLU\\
		\hline
		$512$ & FC, BN, l-ReLU\\
		\hline
		$4 \times 4 \times 32$ & Unflatten  \\
		\hline
		$11 \times 11 \times 64$ & ConvT (st=2), BN, l-ReLU\\
		\hline
		$25 \times 25 \times 128$ & ConvT (st=2), BN, l-ReLU\\
		\hline
		$28 \times 28 \times 1$ & ConvT (st=1), Sigmoid\\
		\hline
	\end{tabular}	
	\quad
	\begin{tabular}[t]{|c|c|}
		\hline
		\multicolumn{2}{c}{Discriminator} \\
		\hline
		Size & Layer\\
		\hline \hline
		$28 \times 28 \times 1$ & Input  \\
		\hline
		$14 \times 14 \times 64$ & Conv (st=2), l-ReLU\\
		\hline
		$7 \times 7 \times 128$ & Conv (st=2), l-ReLU\\
		\hline
		$4 \times 4 \times 256$ & Conv (st=2), l-ReLU\\
		\hline
		$4096$ & Flatten\\
		\hline
		$1$ & FC\\
		\hline
	\end{tabular}
\end{table}

\begin{table}
	\centering
	\caption{Encoder output dimension $dim$, quantization levels $L$, and tradeoff coefficients $\lambda$ used for training the encoder-decoder pairs in the experiments of of Sec.~\ref{sec:exp}}\label{tab:comp_parameters}
	\small
	\begin{tabular}[t]{|c|c|l|}
		\hline
		$dim$ & $L$ & \multicolumn{1}{|c|}{$\lambda$}\\
		\hline \hline
		$2$ & $2$ & $0,2,2.5,3,3.5,4,4.3,4.6,5,5.5,6,8,10,15,20$ \\
		\hline
		$3$ & $2$ & $0,2,2.5,3,3.5,4,4.3,4.6,5,6,8,9,10$ \\
		\hline
		$4$ & $2$ & $0,2,2.5,3,3.5,4,4.3,4.6,4.8,4.9,5,6,8,10$ \\
		\hline
		$4$ & $3$ & $0,2,2.5,3,3.5,4,4.3,4.6,5,6,8,10$ \\
		\hline
		$4$ & $4$ & $0,2,2.5,3,3.5,4,4.3,4.6,5,6,8,10$ \\
		\hline
		$4$ & $6$ & $0,2,2.5,3,3.5,4,5,10$ \\
		\hline
		$4$ & $8$ & $0,2,2.5,3,4,5,6,10$ \\
		\hline
		$4$ & $11$ & $0,2,2.5,3,4,5,6,10$ \\
		\hline
		$4$ & $16$ & $0,2,2.5,3,4,5,6,10$ \\
		\hline
	\end{tabular}
\end{table}

The quantization layer (last encoder layer) follows Mentzer et al.~\yrcite{mentzer2018conditional}. Here, the bin centers $\mathcal{C}=\{c_1,\ldots,c_L\}$ are fixed and evenly spaced in the interval $[-1,1]$. Denoting by $z_i$ the output of the encoder unit $i$ before quantization (after the Tanh activation), the encoder output $\hat{z}_i$ in the forward pass is given by nearest-neighbor assignment, i.e.~$\hat{z}_i = \arg \min_{c_j} \|z_i - c_j \|$. To compute the gradients in the backward pass, we use a differential ``soft'' assignment
\begin{equation}
\tilde{z}_i = \sum_{j=1}^L \frac{\exp(-\sigma \|z_i-c_j \|_1)}{\sum_{l=1}^L \exp (-\sigma \|z_i-c_l \|_1)} c_j,
\end{equation}
where we use $\sigma = 2/L$. Uniformly distributed noise $\mathcal{U}(-\tfrac{a}{2},\tfrac{a}{2})$ is added to the encoder output before it is passed on to the decoder, with $a = 2 / (L-1)$.

\section{The perceptual loss in the experiment in Sec.~\ref{sec:exp_perceptual}}
\begin{figure}[h]
	\vskip 0.2in
	\centering
	\subfigure[]{%
		\includegraphics[width=.12\textwidth]{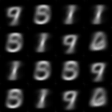} \label{fig:first_layer_perc} 
	} 
	\quad \quad
	\subfigure[]{%
		\includegraphics[width=.12\textwidth]{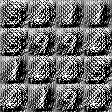}
		\label{fig:second_layer_perc} 
	} 
	\quad \quad
	\subfigure[]{%
		\includegraphics[width=.12\textwidth]{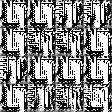}
		\label{fig:third_layer_perc} 
	} 
	\caption{\textbf{Minimizing the perceptual loss alone (without an additional MSE term).} The perceptual loss is evaluated on the outputs of: (a) the first convolutional layer, (b) the second convolutional layer, and (c) the first fully-connected layer.}
	\label{fig:compare_layers_perc}
\end{figure}

\begin{figure}[h]
	\vskip 0.2in
	\centering
	\subfigure[]{%
		\includegraphics[width=.12\textwidth]{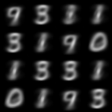} \label{fig:MSE} 
	} 
	\quad \quad
	\subfigure[]{%
		\includegraphics[width=.12\textwidth]{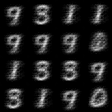} \label{fig:first_layer} 
	} 
	\quad \quad
	\subfigure[]{%
		\includegraphics[width=.12\textwidth]{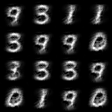}
		\label{fig:second_layer} 
	} 
	\quad \quad
	\subfigure[]{%
		\includegraphics[width=.12\textwidth]{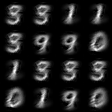}
		\label{fig:third_layer} 
	} 
	\caption{\textbf{Visual comparison of assessing the perceptual loss on the outputs of different layers.} (a) Minimizing the MSE alone. (b)-(d) Minimizing a combination of the MSE and percpetual loss, where the perceptual loss is evaluated on the outputs of: (b) the first convolutional layer, (c) the second convolutional layer, and (d) the first fully-connected layer. The weights of each term (MSE, perceptual) in the loss were optimized for visual quality.}
	\label{fig:compare_layers}
\end{figure}

In the experiment of Sec.~\ref{sec:exp_perceptual}, the distortion term of the optimization objective \eqref{eq:objective} is taken as a combination of the squared-error and the perceptual loss of Johnson et al.~\yrcite{johnson2016perceptual}. We use this combination since minimizing the perceptual loss alone does not lead to pleasing results (Fig.~\ref{fig:compare_layers_perc}), and is commonly used in combination with an additional distortion term, e.g.~$\ell_2\slash \ell_1\slash$contextual loss \cite{ledig2017photo,mechrez2018learning,mechrez2018contextual,liu2018image,wang2018esrgan,shama2018adversarial,shoshan2018dynamic}. As shown in \eqref{eq:perc_loss}, this perceptual loss is in essence the squared-error in \emph{the deep-feature space} of a pre-trained convolutional net. The standard pre-trained net used with the perceptual loss is the VGG net \cite{simonyan2014very}, which is trained on natural images from the ImageNet dataset, and is not appropriate for assessing the similarity between MNIST digit images. We therefore pre-train a simple net for classifying MNIST digit images, which achieves over $99\%$ accuracy. The architecture of this pre-trained net is presented in Table \ref{tab:MNIST_architecture}. The perceptual loss in our experiment is taken as the MSE on the outputs of the second convolutional layer, as this leads to the best perceptual quality (see Fig.~\ref{fig:compare_layers}). We trained with stochastic gradient descent for $30$ epochs with a batch size of $30$. The learning rate was initialized to $10^{-2}$ and decreased by $\tfrac{1}{5}$ after 20 epochs.

\begin{table}
	\centering
	\caption{Architecture of the pre-trained MNIST digit classification net used with the perceptual loss of Johnson et al.~\yrcite{johnson2016perceptual} in the experiment in Sec.~\ref{sec:exp_perceptual}. FC is a fully-connected layer, Conv is a convolutional layer with k denoting the kernel size, MP is a max-pooling layer with w denoting the window size over which the maximum is taken.}\label{tab:MNIST_architecture}
	\small
	\begin{tabular}[t]{|c|c|}
		\hline
		Size & Layer\\
		\hline \hline
		$28 \times 28 \times 1$ & Input  \\
		\hline
		$12 \times 12 \times 10$ & Conv (k=5), MP (w=2), l-ReLU\\
		\hline
		$4 \times 4 \times 20$ & Conv (k=5), Dropout, MP (w=2), l-ReLU\\
		\hline
		$320$ & Flatten\\
		\hline
		$50$ & FC, ReLU, Dropout\\
		\hline
		$10$ & FC, Softmax\\
		\hline
	\end{tabular}
\end{table}

\end{document}